\documentclass{article}
\usepackage{graphicx} 
\usepackage{tabularx}

\usepackage{wrapfig}
\usepackage{url}
\usepackage{booktabs}
\usepackage[table,xcdraw]{xcolor}
\usepackage{enumerate}
\usepackage{amsmath}

\newenvironment{acks}{\section*{Acknowledgments}}{}
\DeclareRobustCommand{\name}{\textsc{SimCoachCorpus}}
\usepackage[table]{xcolor}
\usepackage{listings}
\usepackage{booktabs}
\usepackage{paralist}
\usepackage{float}
\definecolor{darkgreen}{rgb}{0.0, 0.5, 0.0}
\usepackage[utf8]{inputenc}

\title{\name{}: A naturalistic dataset with language and trajectories for embodied teaching}
\author{%
\resizebox{0.95\textwidth}{!}{\begin{tabular}[t]{@{}c@{\hspace{2em}}c@{\hspace{2em}}c@{}}
  Emily Sumner\thanks{Equal contributions} & Deepak E. Gopinath$^*$ & Xiongyi Cui \\
  Toyota Research Institute & Toyota Research Institute & Toyota Research Institute \\
  Cambridge, MA 02139 & Cambridge, MA 02139 & Cambridge, MA 02139 \\
  \texttt{emily.sumner@tri.global} & \texttt{deepak.gopinath@tri.global} & \texttt{xiongyi.cui@tri.global} \\[1.5em]
  Laporsha Dees & Patricio Reyes Gomez & Andrew Silva \\
  Toyota Research Institute & Toyota Research Institute & Toyota Research Institute \\
  Cambridge, MA 02139 & Cambridge, MA 02139 & Cambridge, MA 02139 \\
  \texttt{ltdees00@gmail.com} & \texttt{patricio.reyesgomez.ctr@tri.global} & \texttt{asilvaa9@gmail.com} \\[1.5em]
  Jean Costa & Allison Morgan & Mariah Schrum \\
  Toyota Research Institute & Toyota Research Institute & Toyota Research Institute \\
  Los Altos, CA 94022 & Los Altos, CA 94022 & Los Altos, CA 94022 \\
  \texttt{jean.costa@tri.global} & \texttt{allison.morgan@tri.global} & \texttt{mariah.schrum@tri.global} \\[1.5em]
  Tiffany L. Chen & Avinash Balachandran & Guy Rosman \\
  Toyota Research Institute & Toyota Research Institute & Toyota Research Institute \\
  Los Altos, CA 94022 & Los Altos, CA 94022 & Cambridge, MA 02139 \\
  \texttt{tiffany.chen@tri.global} & \texttt{avinash.balachandran@tri.global} & \texttt{guy.rosman@tri.global}
\end{tabular}}
}

\date{}

\begin{document}

\maketitle

\begin{abstract}
High-quality curated datasets are essential for training and evaluating AI approaches, but are often lacking in embodied interactive domains where language and physical action are deeply intertwined. In particular, few datasets capture how people acquire motor skills in embodied tasks through verbal instruction over time. To address this gap, we introduce \name{}: a unique dataset of race car simulator driving that enables the investigation of rich interactive phenomena during guided and unguided motor skill acquisition. In this dataset, 29 humans were asked to drive in a driving simulator around a race track for approximately ninety minutes. Fifteen participants received personalized one-on-one instruction from a professional performance driving coach, and 14 participants drove without coaching instruction. \name{} includes features such as vehicle state and inputs, map (track boundaries and race-line), and cone landmarks.  Additionally, these are synchronized with the coach's concurrent verbal feedback and additional terminal feedback at the end of each lap. We also provide high-quality annotations of high-level coaching categories for each concurrent feedback utterance, ratings on students' compliance with coaching advice, and self-reported cognitive load and emotional state of participants (gathered from surveys during the study). The final dataset includes over 20,000 concurrent feedback utterances, over 400 terminal feedback utterances, and over 40 hours of interactive driving data. Our naturalistic interactive dataset can be used to investigate motor learning dynamics, explore linguistic phenomena, and train computational models of teaching and learning. We demonstrate applications of this dataset for in-context learning, imitation learning, and topic modeling. Data is hosted at \url{https://doi.org/10.7910/DVN/W7VTKZ}  and code is available at \url{https://github.com/ToyotaResearchInstitute/sim_coach_corpus}.
\end{abstract}

\section{Introduction}
The field of education has been significantly affected by artificial intelligence (AI)~\cite{bieletzke2024smarta,Khan_2023,zhai2021review,wang2024artificial}. It has been shown that individualized tutoring leads to significantly better learning outcomes~\cite{bloom19842}. AI has the potential to enable personalized tutoring at scale and across contexts. However, these systems are limited by the data that they are trained on~\cite{sambasivan2021everyone}. 

Data for teaching over chat messages ~\cite{caines2020teacher,caines2022teacher,wang2024tutor},  classroom-level instruction ~\cite{martinez2020moodoo}, and long-term educational outcomes ~\cite{chatterji2006reading, halle2009disparities} has been relatively available. However, datasets in educational domains with physical or motor aspects are scarce, especially when searching for longitudinal data. These limitations hamper both the development of AI systems for embodied learning and limit the reproducibility of the resulting research in domains where embodied expertise is central. These challenges arise in areas such as sports education and high-performance driving, where verbal instructions and physical performance are tightly coupled, but also extend to broader contexts, including early childhood education, musical training, rehabilitation, and surgical residency. 
 
Motorsports is an appealing domain for motor learning datasets because of its well-defined performance measures, physical features that indicate students' behavior, and relatively rapid improvement for novices. While motorsports datasets have been collected in prior work \cite{remonda2024simulation}, no dataset tracks students longitudinally, or includes embodied student data, longitudinal interaction, performance, student state, and in-person one-on-one tutoring. 

\textbf{Contribution}: In this paper, we introduce \name{}, a unique dataset of interactive coaching in the high-performance driving education (HPDE) domain that aims to reduce the above-mentioned gaps by capturing longitudinal interactive motor learning phenomena in the presence of expert instruction. Our dataset offers unique opportunities for researchers from various fields to explore key research questions such as a) how language adapts over time with or without conditioning on embodied state information and b) how student compliance to instructions mediates learning. It also allows for the development of machine learning systems for coaching; and naturalistic linguistic phenomena and their relationship to student behavior. The dataset includes: i) personalized one-on-one verbal instruction during and after driving with the same expert coach, collected in a longitudinal learning context;
ii) synchronized dense state and action information with expert teaching utterances and annotations, along with metadata on the students' emotional state and cognitive load; 
iii) Self-practice (no coaching) control group to understand the influence of coaching on both student behavior and overall performance; iv)  Example code for several relevant machine learning tasks, such as behavior cloning of coach and student, as well as text analysis and generation.

\name{} is the first dataset to jointly capture longitudinal motor learning, expert language, and dense embodied state under controlled instruction. It facilitates exploring questions such as how instruction timing affects motor learning, how compliance mediates learning, and how language adapts over time with or without relation to the embodied student state. 
\section{Background: existing datasets}

Our dataset relates to those in several fields of research -- namely, educational datasets, driving datasets, and task-specific multimodal datasets. While there are additional datasets involving linguistic contents \cite{liu2024datasets}, in this section we focus on those most closely aligned with the gap our work addresses---longitudinal, embodied instruction, teaching, with high-performance driving as a domain of focus.

\textbf{Educational Datasets}
Within the field of AI in education \cite{chen2020artificial}, and in subfields such as knowledge tracing~\cite{abdelrahman2023knowledge}, significant attention has been devoted to the development of relevant datasets.
Teacher interaction datasets either explore conversations~\cite{settles2018second,Suresh22}, spatial behavior~\cite{martinez2020moodoo}, or video recordings for behavioral analysis~\cite{sun2021student}. Some contain repeated interactions, but do not include embodied state features ~\cite{caines2020teacher,caines2022teacher}. Similarly, most longitudinal datasets~\cite{chatterji2006reading,konstantopoulos2023class,shangguan2024scalable} focus on high-level features, rather than capturing embodied or fine-grained observations of the instruction process.
Additional instructional language datasets have been collected with an aim to improve LLM training~\cite{zhang2024sciinstruct} but do not focus on student-instructor dynamics.
Finally, another class of datasets appears in specific domains where embodiment is key, and education is prominent in the community, such as surgical societies~\cite{rojas2020daisi}. However, these too are often limited in terms of longitudinal span or ecological validity.

\textbf{Sports Datasets}
Our dataset is also related to sports analysis datasets, \cite{ma2025t3set,zhao2025survey}, although most existing work focuses on specific recognition, tracking, and prediction tasks related to the specific sport, rather than the higher-level analyses of the athlete's skill development or the instructional process. In cases where sport coaching is involved, datasets are often either based on high-level features~\cite{Lorimer09}, linguistic aspects~\cite{xia2024language}, or other summary statistics. Datasets that include both raw instruction video and linguistic information~\cite{10203838} still often lack longitudinal information or performance metrics. \cite{ma2025t3set} uniquely contains coach-student dyads coupled with sensor data. They introduce a coaching dataset for table tennis, both concurrent and terminal feedback are present; however the coach's instructions were collected retrospectively and therefore do not contain true interaction with the student. Additionally, the longitudinal component of student learning and coach adaptation is missing as well.  

\textbf{Driving Datasets}
Our dataset also relates to driving datasets, especially those of motor racing~\cite{Kulkarni_2023,remonda2024simulation}. Many of the driving datasets focus on vehicle perception or behavior modeling \cite{geiger2013vision,kim2018textual,ettinger2021large,caesar2020nuscenes,gulino2023waymax,caesar2021nuplan,sun2020scalability} and do not explore characteristics of the driver. Even more human-centric ones \cite{Gopinath_2021_ICCV,alletto2016dr,yuen2016looking,rasouli2019pie} rarely probe how drivers are coached during driving, or how instruction is personalized.
There are a few examples that probe coaching in driving, but do not provide a full multimodal dataset with both naturalistic interaction and full vehicle states and controls~\cite{srivastava2025shared,schrum2025skill}.

\section{Dataset overview}
This section presents an overview of the data acquisition process, study design and recruitment pool.
\subsection{Data acquisition}

We collected data in a static racing simulation environment, based on a CARLA simulation engine \cite{dosovitskiy2017carla}, adapted to track racing (see \cite{gopinath2024computational} for details). Student participants were split into two conditions: a coached condition and a self-practice condition. Participants were recruited through two recruiting agencies, Fieldwork Boston and User Interviews. Participants signed a consent form and; they were compensated \$150 for their participation. This work was reviewed and approved by the WCG IRB (Protocol Number 20232162).

\begin{figure*}
    \centering
    \includegraphics[width=\linewidth]{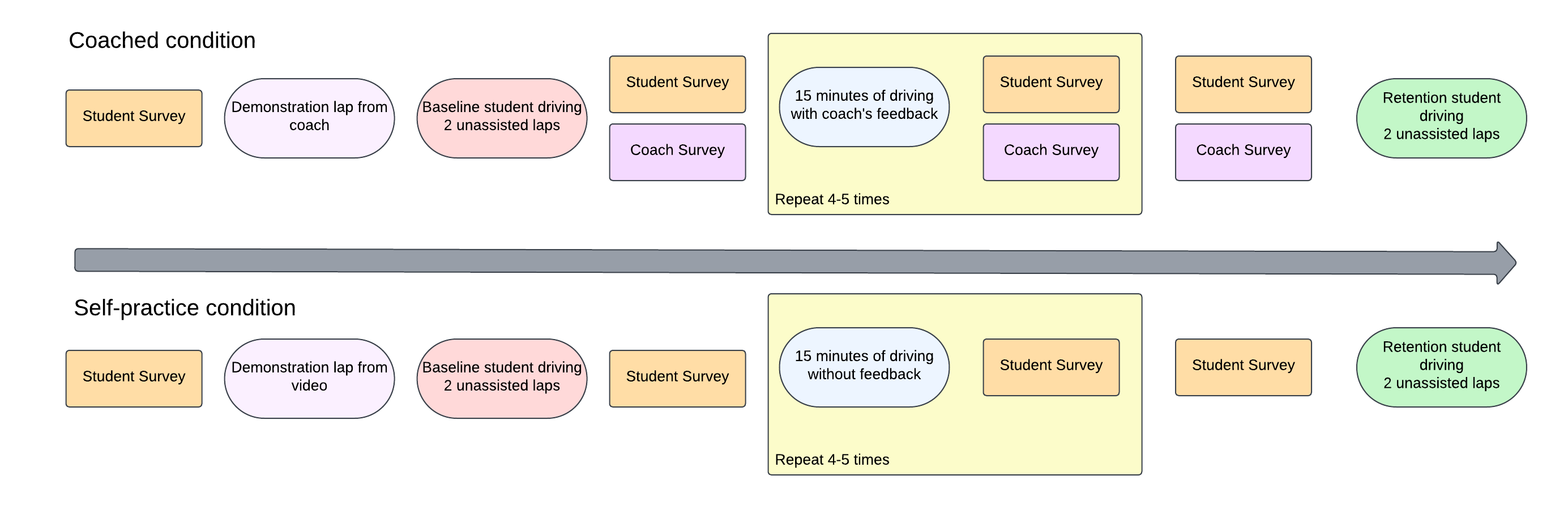}
    \caption{Overview of the study structure, for both self-practice and coached participants.}
    \label{fig:study-overview}
\end{figure*}

\begin{figure}
    \centering

 \begin{minipage}{0.3\textwidth}
        \centering
        \includegraphics[width=0.9\linewidth]{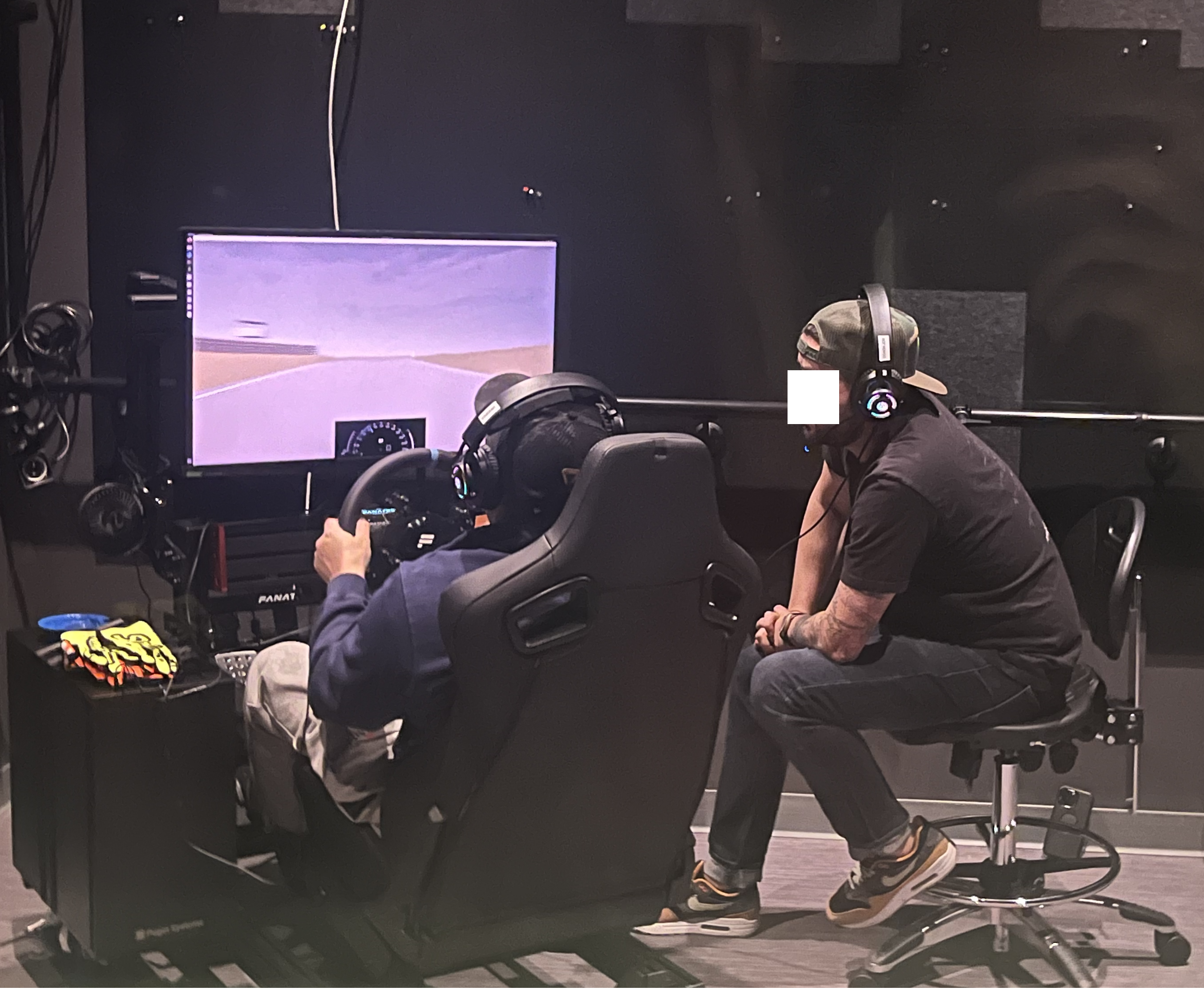}

    \end{minipage}    
        \begin{minipage}{0.45\textwidth}
        \centering
        \includegraphics[width=0.9\linewidth]{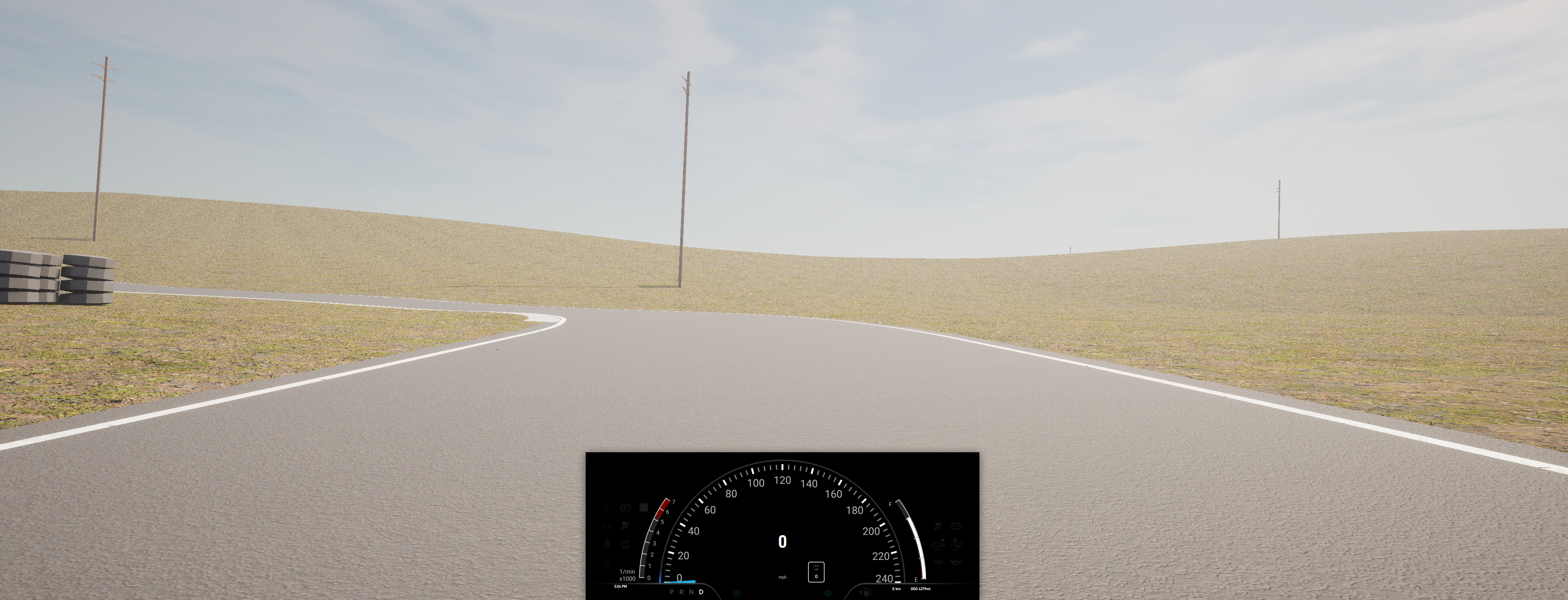}

    \end{minipage}    

    \caption{Top: Study setup for coached participants.  Bottom: Participants' field of view while driving in the simulator.}
    \label{fig:study-setup}
\end{figure}

\subsubsection{Study structure}
Figure \ref{fig:study-overview} shows an overview of the study structure. Participants first filled out a consent form that outlined the risks of the study (eye strain \& motion sickness). After completing this form, participants filled out demographic surveys. Afterwards, participants entered the simulator room. Those who were in the coaching condition met the coach at that time. 

Before driving, students were shown a demonstration lap that explained high-performance driving fundamentals. This demonstration explained braking zones and apexes. For coached participants, this demonstration lap was driven by the coach while the student watched. For self-practice participants, a video was presented on the simulator as the wheel moved, providing the student with haptic feedback, along with a bargraph of the brake and throttle inputs. See \url{https://youtu.be/PgLtIH5Qbv0} for the video \& audio projected on the screen. After the demonstration lap, participants in both conditions drove two laps as their baseline performance laps, which we refer to as \textit{Baseline} laps. The coach provided no feedback to participants in the coaching condition during the baseline laps. 

Participants in both conditions drove in 15-minute blocks and at the end of each block were then prompted to take surveys assessing their cognitive load via the NASA TLX~\cite{hart1988development}, emotional state via the PANAS \cite{crawford2004positive}, amount of fun (scale of 1-10), and verbally reported their level of motion sickness via the Fast Motion Sickness Scale~\cite{keshavarz2011validating}. If participants experienced a large increase in their motion sickness score, they were offered the options of beverages and ginger candy, taking a break, or stopping the study early. One participant was unable to complete 2 laps due to motion sickness.

In the coached condition, participants worked with a professional performance driving coach during the 15-minute driving blocks. All participants were coached by the same expert to ensure cohesion across instructions. The coach provided two types of feedback: \emph{concurrent feedback} and \emph{terminal feedback}. Concurrent feedback refers to the coaching the participant received while driving. Terminal feedback refers to the conversational feedback provided at the end of each lap.  In order to preserve naturalism, it was up to the coach's full discretion whether to give terminal feedback or not. The coach gave terminal feedback 75.92\% of the time. At the end of the study, all participants completed two \textit{Retention} laps without coaching. Appendix~\ref{appendix:questionnaires} points to details on the questionnaires.

\subsubsection{Participant demographics}
\begin{table}
\caption{Recruited participant demographics.}
\label{tab:demographics}
\centering
\footnotesize
\begin{tabular}{lll}
\toprule
 & Self-Practice ($n=15$) & Coached ($n=15$) \\
\midrule
Age (mean $\pm$ SD) & 39.3 $\pm$ 13.6 & 36.9 $\pm$ 9.7 \\
Range & 23--63 & 23--54 \\
Female / Male & 7 / 8 & 7 / 8 \\
White & 11 & 3 \\
Latinx or Hispanic & 2 & 4 \\
Black or African American & 1 & 3 \\
Asian & 1 & 5 \\
Excluded (motion sickness)  & 1 & 0 \\
\bottomrule
\end{tabular}
\end{table}

Table ~\ref{tab:demographics} shows the demographic information of the recruited 30 participants. Note that only 29 were included in the analysis as one participant dropped out due to motion sickness early on. Our expert instructor has been doing professional performance driving for 10 years, coaching for over six years, with six years of experience coaching on track and six years of experience coaching in simulators.

\section{Dataset statistics}
\name{} captures several aspects of the training and coaching process. It includes densely sampled vehicle state and control information, the map and environment state, and the verbal interactions between the student and the coach in the coached condition. Our data also includes student-reported affect and cognitive load collected at 15-minute intervals. We collected affect and cognitive load as which have been shown to be related to student learning outcomes and experience \cite{munchow2017better, lodej2025investigating, costa2025dashboards,d2012dynamics}. In this section, we demonstrate the richness of this dataset with some descriptive statistics and analysis.

\subsection{Vehicle trajectory statistics}
\label{subsec:veh_traj_stats}
\begin{figure}
 \begin{minipage}{0.5\textwidth}
        \centering
        \includegraphics[width=0.95\linewidth]{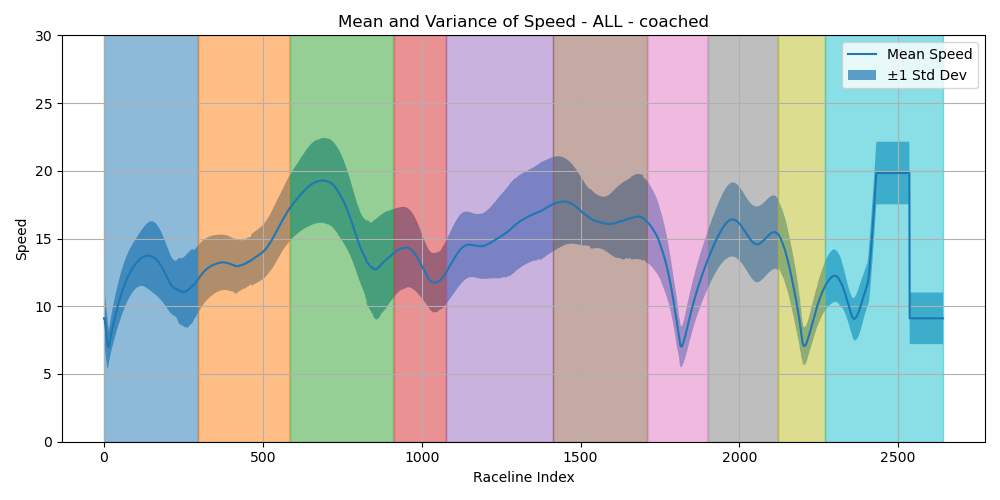}
    \end{minipage}    
        \begin{minipage}{0.5\textwidth}
        \centering
        \includegraphics[width=0.95\linewidth]{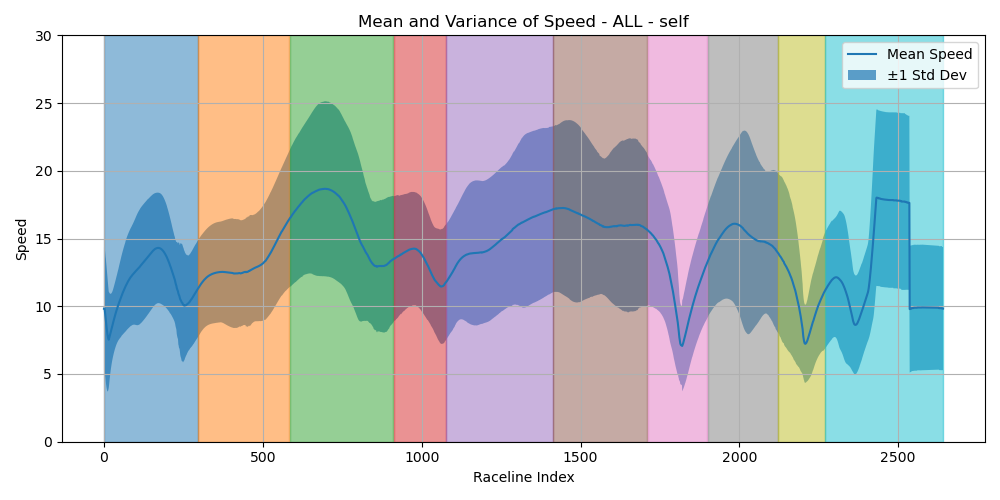}
    \end{minipage} 
    \begin{minipage}{\textwidth}
        \centering
        \includegraphics[width=0.45\linewidth]{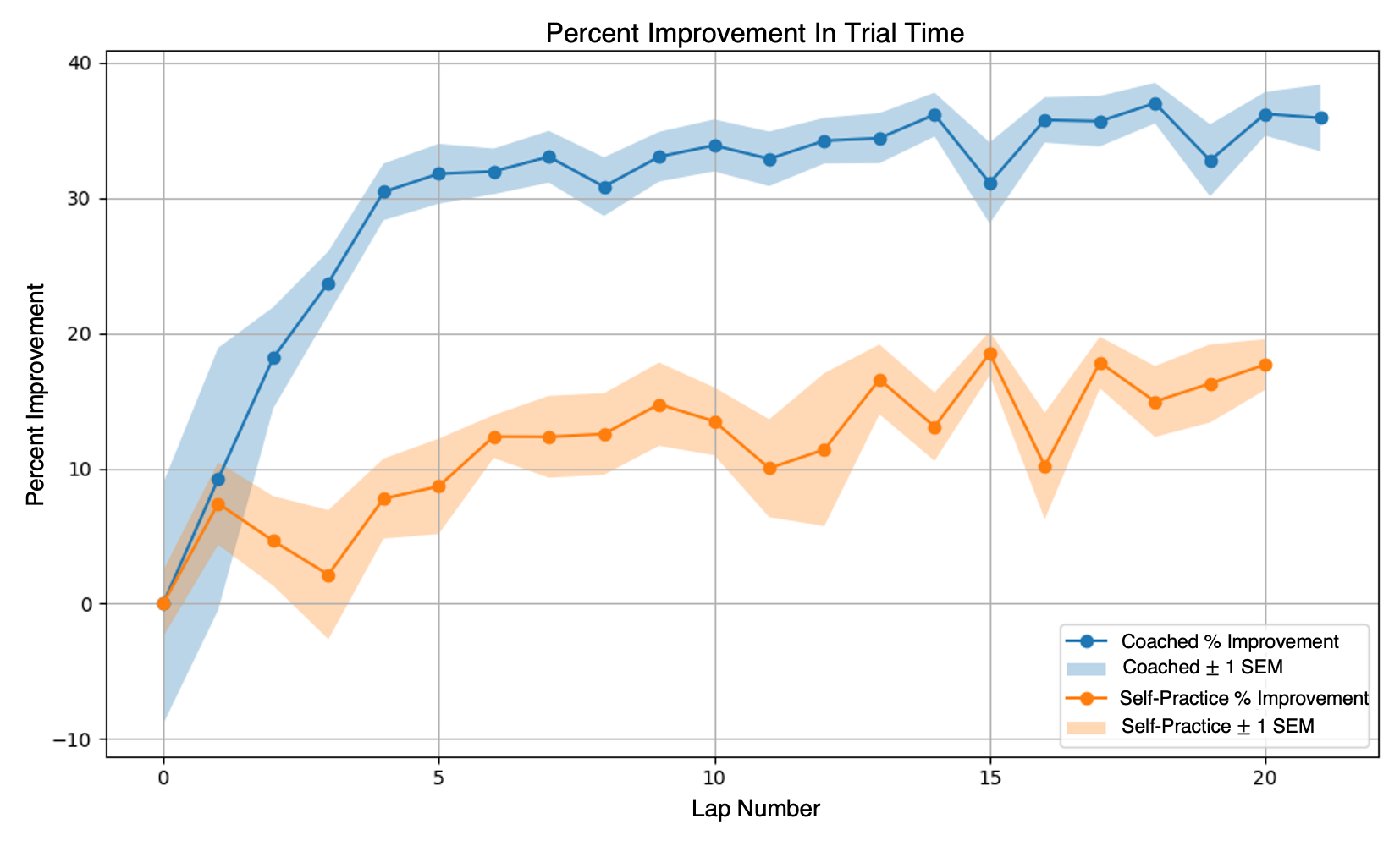}

    \end{minipage}
    \caption{Mean and variance of speed across subjects for coaching (left) and the self-practice (middle) as a function of longitudinal track position. There is a lot more variance in the self-practice condition. The colored bands represent the different track segments. Comparison of percent improvement over time in coached vs self-practice participants with +/- 1 standard error ribbons (right).}
    \label{fig:traj_behavior_speed}
\end{figure}

We analyzed how driving behavioral features such as speed, steering, and throttle control profiles change as a function of location on the track as well as across subjects. Figure~\ref{fig:traj_behavior_speed} depicts the mean and variance of the speed profile across participants for both the coached and the self-practice datasets. We observe that, under the coached condition, the variance of speed is much lower compared to the self-practice condition suggesting that with guidance subjects are able to more consistently control the vehicle.  To better understand behavior, we divided the track into discrete segments (i.e., track-segments), each representing a functionally distinct portion of the course (e.g., straightaways, gentle curves, and sharp turns). We observe track-segment-specific characteristics, for instance, the lowest speed always happens at the sharpest hairpin turn. Appendix \ref{appendix:additional_coached_self-practice} points to additional behavioral plots related to steering and throttle control profiles.

\begin{table}[h]
  \caption{Trajectory metrics: comparison of Coached vs Self-Practice across trial types. For all metrics, lower is better.}
  \label{tab:self-v-coached-metrics}
  \centering
  \resizebox{0.9\textwidth}{!}{%
  \begin{tabular}{lcccccc}
    \toprule
    Trial Type     & \multicolumn{2}{c}{Lap Time (s)$\downarrow$} & \multicolumn{2}{c}{Out-of-Bounds (\%)$\downarrow$} & \multicolumn{2}{c}{Raceline Adherence (m)$\downarrow$} \\
                   & Coached         & Self            & Coached         & Self              & Coached         & Self             \\
    \midrule
    Baseline       & 145.00 (10.30)  & 125.00 (7.50)   & 11.50 (1.17)     & 14.60 (2.32)     & 2.50 (0.08)     & 2.97 (0.30)      \\
    Driving laps   & 101.00 (0.78)   & 99.30 (0.63)    & 10.80 (0.38)     & 16.10 (0.46)     & 2.46 (0.04)     & 3.01 (0.06)      \\
    Retention      & 100.00 (2.23)   & 96.40 (2.38)    & 10.30 (1.29)     & 17.10 (1.83)     & 2.32 (0.12)     & 2.92 (0.22)      \\
    \bottomrule
  \end{tabular}
  }
\end{table}
\subsection{Coached vs self-practice performance}
\label{subsec:coached_self_practice}
Table~\ref{tab:self-v-coached-metrics} shows the differences in Baseline, Retention, and the Driving laps between the two different conditions. Figure~\ref{fig:traj_behavior_speed} shows differences between the coached and self-practice participants. To explore significant difference, we fit a series of linear mixed models predicting various metrics with condition, lap number, and their interaction as fixed effects for lap number by participant (\textit{metric} $\sim$ condition $\times$ lap number + (1 + lap number $|$ participant) using \texttt{lmerTest}~\cite{kuznetsova2017lmertest} in R. All $\beta$ estimates and $p$-values reported in this subsection are from these models. Percent improvement is calculated as the percent decrease in lap time compared to the initial baseline lap. 

Notably, coached participants achieved significantly more than a 35\% improvement compared to their initial lap times, whereas self-practice participants achieved less than a 20\% improvement $\beta = 24.24, p = 0.006$), despite initial slower laps, which may be due to phenomena such as~\cite{bjork2013desirable,shea1979contextual}. 

Building on prior work \cite{schrum2025skill}, we also examined domain-relevant skill metrics to gain deeper insight into underlying ability. Specifically, we analyzed racing line adherence, percent of time out of bounds, and average lateral g-force during cornering. Unlike lap time, which is an outcome influenced by many factors, these skill metrics capture underlying driver ability, which is more indicative of long-term skill development. 

Additionally, across all driving trials, coached participants had significantly better raceline adherence than participants in the self-practice group (linear mixed-model, $\beta = 0.82, p = .014$). Coached participants also spent significantly less time out-of-bounds than self-practice participants across all trials (linear mixed model, $\beta = 7.65, p = 0.013$). While not statistically significant, we found that coached participants improved their average g-force during cornering by over 40\% whereas self-practice participants only improved by 30\%. This suggests that coaching could help people achieve greater performance than self-practice, where people are unguided and unbounded in what they can try out. See Appendix~\ref{appendix:additional_coached_self-practice} for additional results on self-practiced vs. coached conditions.

\subsection{Student state and performance}
We ran a linear mixed-effects model using the student-state data from the NASA TLX and PANAS to characterize mean lap time per session with fixed effects of session, condition, how much fun participants were having, positive \& negative affect, perceived effort, perceived physical demand, feeling of success, and perception of pacing, with a random intercept for participant. This was done using the lmerTest package in R \cite{kuznetsova2017lmertest}, p-values are calculated using Satterthwaite's method. There was a significant effect of session number ($b = -2.60, SE = 0.53, t(94.44) = -4.92, p < .001$), and how much fun participants were having ($b = -2.27, SE = 0.94, t(100.2) = -2.41, p = .018$). There was a marginal effect of the self-practice group being faster ($b = -6.66, SE = 3.87, t(29.05) = -1.72, p = .096$), which can be interpreted as mentioned in Subsection~\ref{subsec:coached_self_practice}. 

Table~\ref{tab:lmm-student-state-lap-time} shows more details about the linear mixed-effects model run to explore the relationship between lap time and student state. Notably, session number and how much fun a student is having are significant predictors.

\begin{table}[ht]
\centering
\caption{Linear Mixed Model Predicting Session Lap Time Mean}
\label{tab:lmm-student-state-lap-time}
\small
\begin{tabular}{lrrrr}
\toprule
\textbf{Predictor} & \textbf{$\beta$} & \textbf{SE} & \textbf{\textit{t}} & \textbf{\textit{p}} \\
\midrule
Intercept              & 140.70 & 11.24 & 12.51  & $<$.001 *** \\
Session                & -2.60  & 0.53  & -4.92  & $<$.001 *** \\
Condition (Self-Prac.) & -6.66  & 3.87  & -1.72  & .096 *      \\
Fun                    & -2.27  & 0.94  & -2.41  & .018 **     \\
PANAS Positive         & -0.35  & 0.26  & -1.35  & .179        \\
PANAS Negative         & -1.18  & 0.91  & -1.30  & .195        \\
Effort                 & -0.07  & 0.26  & -0.29  & .777        \\
Physical Demand        & 0.28   & 0.43  & 0.66   & .511        \\
Success                & 0.00   & 0.29  & 0.00   & .998        \\
Pacing                 & 0.51   & 0.35  & 1.45   & .151        \\
\bottomrule
\end{tabular}
\begin{flushleft}
\footnotesize{\textit{Note.} * $p < .10$, ** $p < .05$, *** $p < .001$. Random intercept variance = 83.84; residual = 38.85. REML = 821.7.}
\end{flushleft}
\end{table}

Wilcoxon signed-sum tests revealed that participants in the coached group had significantly higher ratings on both the fun metric ($W = 2671, p < .001$) and the PANAS positive emotions score ($W = 2674.5, p< .001$) than participants in the self-practice condition. No significant differences appear on the PANAS negative emotions score or any of the NASA TLX cognitive load metrics. This suggests that coaching may positively influence student state. Further exploration of this, both within the dataset and through future studies, is warranted.  

\subsection{Concurrent feedback}\label{statistics:cf}
 We analyze the contents of the concurrent feedback in terms of both the topic of instruction provided and the temporal changes in topic distribution over time. 

Across all 15 coached participants, there were a total of 21,356 utterances that the coach spoke while participants were driving. 20,938 sentences (98$\%$) had an inter-annotator agreement greater than 60$\%$. All participants completed at least 21 laps within their teaching session with an average of 22.6 ($sd = 4.14$). Our analysis will focus on trends seen within the first 21 laps. Annotated sentences were time-synchronized with the vehicle state and control information for more fine-grained spatial and temporal analysis.

We categorized utterances into four categories:
(i) \textit{instruction} (72.19\%),
(ii) \textit{feedback} (13.54\%),
(iii) \textit{commentary} (9.14\%), and
(iv) \textit{other} (5.13\%).  The \textit{other} category consists primarily of one-word affirmations such as ``Yup'', ``Alright'', or ``Yeah, of course'' in response to a participant's question.  In our analysis, we focus primarily on the \textit{instruction} category. More details on the concurrent feedback annotation procedure are provided in Appendix \ref{appendix:cf-annotation-guidelines}.

The 15,115 utterances labeled by annotators as \textit{instructions} are divided into six main categories that relate to key performance areas required in HPDE: \textbf{throttle} (35.75\%), \textbf{lateral position} (22.24\%), \textbf{steering} (14.44\%), \textbf{looking ahead} (11.90\%), \textbf{turn} (9.82\%), and \textbf{brake} (5.86\%). The instruction categories were derived from conversations with the coach before the study took place. Throttle instructions relate to how much gas is being applied to the throttle (e.g., `more gas'). Lateral position instructions refer to the lateral placement of the vehicle on the track(e.g., `move to the right'). Looking ahead instruction has to do with where the student is looking (e.g., `look at the cone'). When and how much brake pressure is to be applied is captured by brake instructions (e.g., `brake now'). Turning is about explaining when and where to turn (e.g., `now turn in'). Steering-related instruction conveys information regarding the landmarks and direction in which the vehicle should be guided. (e.g., `move closer to that cone'). See Appendix~\ref{appendix:tax-details} for more information on this taxonomy and the annotation process.

\subsubsection{Topic analysis}
In addition to hand-crafting a taxonomy and creating manual annotations for concurrent feedback utterances, we also applied unsupervised topic modeling methods to discover underlying themes in the concurrent feedback. We use a transformer-based topic modeling algorithm, BERTopic~\cite{grootendorst2022bertopic}.

We utilize PCA with three components and K-Means with 20 clusters as our choice of dimensionality reduction and clustering algorithms, respectively. The topic representations learned via BERTopic on the corpus of all concurrent feedback sentences yield interpretable topics that relate to various important aspects of high-performance driving such as braking, throttle control, etc. These topic representations are robust to the choice of embedding models (\texttt{all-MiniLM-L6-v2} and \texttt{all-mpnet-base-v2}). 

\begin{figure}[H]
    \centering
    \includegraphics[width=1\linewidth]{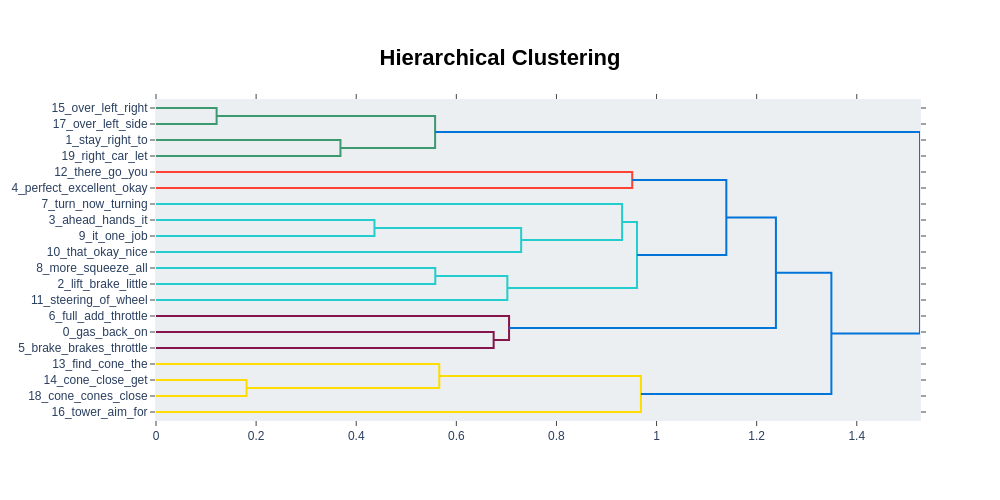}
    \caption{Hierarchical structure of topics discovered. Correlated themes that correspond to important HPDE performance characteristics are clustered together. }
    \label{fig:hierarchical_topics}
\end{figure}

Table~\ref{tab:topic_modeling_table} shows some of the representative topics discovered and example utterances assigned to these topics. Additionally, the unsupervised model was able to achieve a RAND score~\cite{rand1971objective} of 89.1$\%$ and a V-measure score~\cite{rosenberg2007v} of around 50.2$\%$ on a held-out test set. The hierarchical structure of the discovered topics is presented in Figure ~\ref{fig:hierarchical_topics}. We observe that categories related to lateral instructions (``move over to the left'', ``stay right to'') are clustered together. Likewise, topics related to ``looking ahead'' at landmarks such as cones and towers get correctly clustered as well. This indicates that the coach's language exhibit strong structure which can be leveraged as data priors in various modeling tasks. 

\begin{table}
  \caption{Top 5 topics (denoted with the top 5 words within the topic) by BERTopic and representative utterances and percentage of utterances assigned to the topic. }
  \label{tab:topic_modeling_table}
  \centering
  \begin{tabular}{p{0.35\columnwidth} p{0.35\columnwidth} p{0.10\columnwidth}}
    \toprule
      Topic & Utterance Example  & Percent \\ \hline
      [gas, back, on, off] &    ``back on the gas''   &        9.8\%             \\ \hline
      [there, you, go, more] &   ``There you go''   &        9.6\%               \\ \hline
      [stay, right, to, side] &  ``stay to the right''     &     7.6\%                  \\ \hline
      [lift, brake, little, bit] &  ``lift, lift, lift''    &      7.2\%                 \\ \hline
      [ahead, hands, it, look]&   ``keep looking ahead''    &      6.7\%                 \\ \hline
    \bottomrule
  \end{tabular}
\end{table}

\subsubsection{Instruction category distribution by location}
\begin{figure}[H]
\begin{minipage}{0.48\textwidth}
        \centering
        \includegraphics[width=0.9\linewidth]{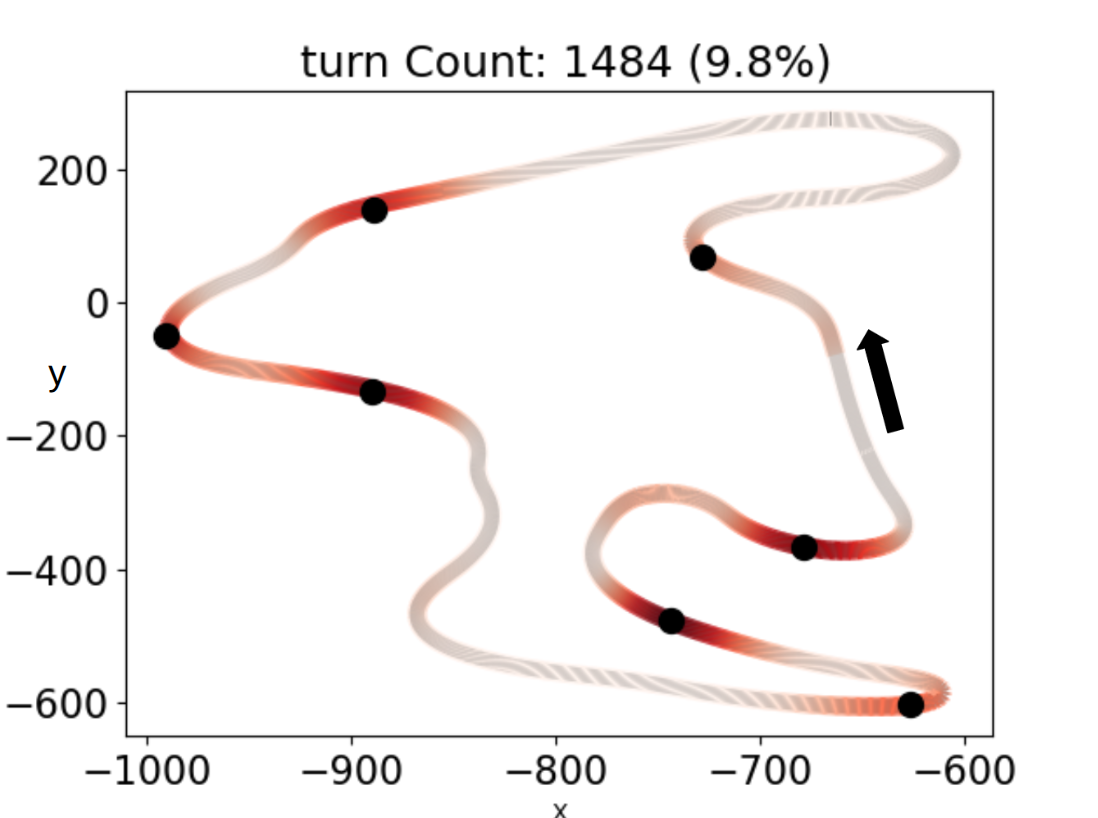}
    \end{minipage} 
 \begin{minipage}{0.48\textwidth}
        \centering
        \includegraphics[width=0.9\linewidth]{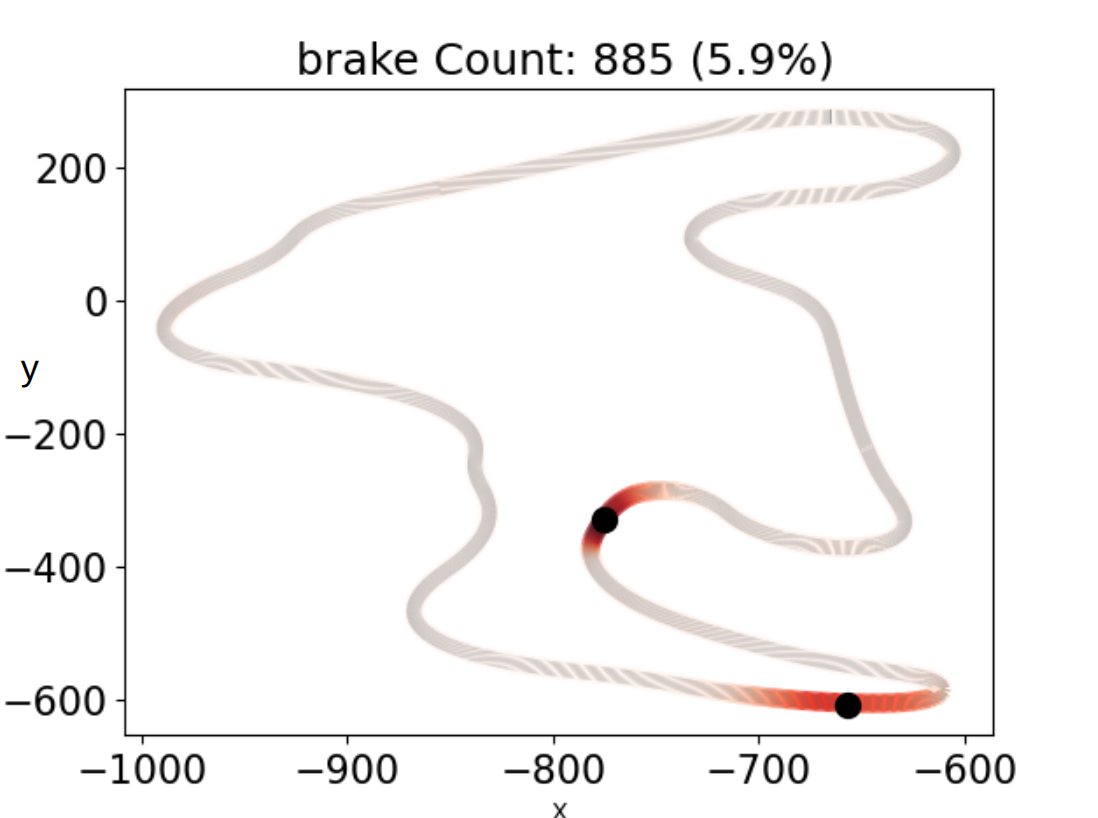}
    \end{minipage}    
       
    \caption{Density of coach utterances along the track for two categories. The black dots in the figure are the discovered modes and exhibit high location specificity. Participants drove counterclockwise, indicated by the arrow in the top figure. Note the location of \textbf{brake} and \textbf{turn} instructions before sharp turns and medium turns, respectively.}
    \label{fig:category-heatmap}
\end{figure}

The geometry of different track locations, to a large extent, determines the distribution of teaching actions. For example, when the car approaches a tight hairpin turn, there is a higher likelihood of the coach issuing brake-related instruction. Figure~\ref{fig:category-heatmap} demonstrates the distribution of two different teaching categories across the track locations. We used one-dimensional kernel density estimation along the longitudinal coordinates of the track with a kernel width of 50 meters to obtain density maps of instructions' spatial distribution. As can be seen, although \textbf{brake} category constitutes less than six percent of all instructions, there is high location specificity. Unsurprisingly, brake-related instructions occur right before the sharpest turns on the map. Similarly, we observe that \textbf{turn}-related instructions tend to clump together right before medium to sharp turns. This correlation between instruction type and context can be exploited by data-driven models to learn context-dependent teaching systems for high performance driving education. 

\subsection{Terminal feedback}

Terminal feedback, the content the coach uttered between each lap, was also analyzed on the types of topics instructed and how those topics changed over time. The coach spoke 411 total terminal feedback sentences. One annotator manually categorized each sentence into eight categories: \textbf{steering} (30.1\%), \textbf{throttle} (17.1\%), \textbf{looking ahead} (14.9\%), \textbf{turn} (14.8\%), \textbf{lateral position} (11.7\%), \textbf{brake} (8.1\%), \textbf{non-instruction} (3.4\%), and \textbf{other} (0.1\%). Appendix~\ref{appendix:cf-annotation-guidelines} gives further definitions for these categories and annotation guidelines.  

The non-instruction category primarily consists of full terminal feedback sentences that have no relation to any type of instruction. An example would be the following: ``Perfect. All right. We're just going to go right into another one. You feel up for it? Perfect.''  In these instances, the coach is providing positive affirmation, checking in on the participant and initiating another lap, which are topics that do not relate to any instruction as that was the primary focus. 

When addressing the instruction categories, the coach primarily does the following: (1) provides advice on how to improve a skill or deepen the participant's understanding of a skill and (2) points out specific actions of a student related to a skill.   

\subsection{Student compliance} 
As an additional label, we annotated student compliance of instructions during the driving sessions. Five annotators went through each utterance paired with an animated rendering of the student's driving session. Each utterance was labeled on whether it was actionable, the category of instruction, how well on a scale of 1-7 the student complied with the instruction (1 being not at all 7 being very well), and how long it took the student to respond to the feedback. More information on the instructions given to the annotators can be found in Appendix~\ref{appendix:comp-annotation}. After removing instances of non-instruction utterances, utterances where it was impossible to tell how well somebody complied (e.g., `looking ahead'), there were a total of 10974 utterances included in these analyses. 

We used a linear mixed-effect model with the lmerTest package in R \cite{kuznetsova2017lmertest} with p-values calculated via Satterthwaite's method to predict compliance ratings based on trial number and instruction category, with random intercepts for each participant. There was a significant main effect of trial number ($b = 0.011, SE = 0.005, t(10960) = 2.434, p = .015$). This suggests that compliance slightly increases over time. Lateral position ($b = -0.283, SE = 0.085, t(10960) = -3.34, p < .001$) and steering ($b = -0.250, SE = 0.091, t(10960) = -2.76, p = .006$) showed significantly lower compliance than braking (the reference category). When looking at relationships between trial and instruction categories, there was a significant interaction between trial number and lateral position ($b = -0.023, SE = 0.005, t(10960) = -4.58, p < .001$). This suggests that there was a lower rate of improvement to following lateral position improvements. Further details on the results of this model can be found in Appendix~\ref{appendix:compliance-analysis-stats}.

\begin{figure}[H]
 \begin{minipage}{\textwidth}
        \centering
        \includegraphics[width=0.45\linewidth]{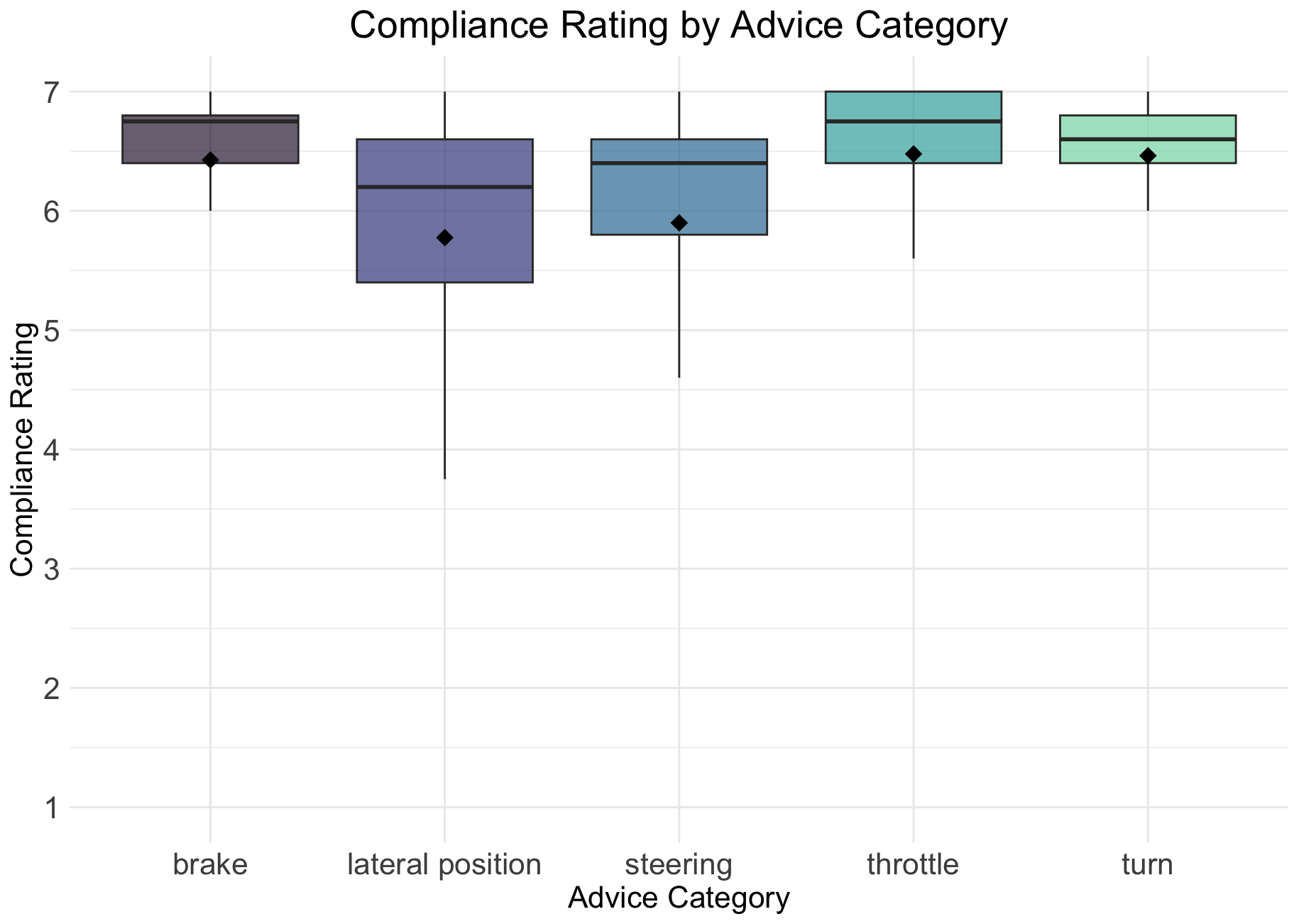}
    \end{minipage}    
    \caption{Boxplots showing the difference in compliance across the different instruction categories (higher is better). Diamonds represent the means. Participants had the lowest compliance when following lateral position instructions. \label{fig:compliance-overview}}
\end{figure}

Figure~\ref{fig:compliance-overview} shows the differences between categories. Lateral control had the lowest compliance. Further, a one-way ANOVA showed significant differences across different instruction categories and compliance, \textit{F}(4, 10969) = 291.40, \allowbreak $p < .001$, $ \eta^2 = .10, 95\%\, \allowbreak \text{CI}\,[.10, 1.00]$  See Appendix~\ref{appendix:compliance-analysis-stats} for additional analysis of compliance.

\section{Example ML usecases}
To demonstrate relevant ML use of the dataset, we have examined two main ML applications: LLM in-context learning for text prediction tasks in the transcripts, and imitation learning for student behavior and teacher action prediction.

\subsection{In-context learning}
The dataset presents several opportunities for language-related machine learning. Here we explore in-context learning (ICL) using off-the-shelf LLMs (GPT-4o~\cite{hurst2024gpt}) to generate terminal feedback for a lap, given concurrent feedback text during the lap as the input. We experimented with adding additional information (lap times and smoothness scores) to the input prompt. 
Figure~\ref{fig:icl_bsd} shows the input prompt, which has the concurrent feedback text and segment-level metrics and the LLM-generated terminal feedback for this prompt. For brevity we don't include the full set of five in-context examples used in the prompt. Additional responses from another SOTA model (Claude Opus 4.7) can be found in the Appendix \ref{appendix:icl_prompts}.

We notice that when provided with the additional embodied information, the LLM is able to ground its response in some factual elements described in the prompt, as shown in Figure~\ref{fig:icl_bsd}, where we observe references to track segments with poor performance. Even with scarce information, the LLM is able to key into specific patterns in the utterances and infer that the student is doing well on specific skills (such as visual anticipation). Going beyond pure language, the dataset allows us to inject state and vehicle control information into the input prompt to improve specificity in feedback.

\begin{figure}[H]
        \centering
    \hfill
    \begin{minipage}{0.49\textwidth}
    \tiny
    GPT Prompt: 
    
    \texttt{
    "P603's trial 11 concurrent feedback contained the following: Little turn to the left. Start lifting off the gas. Look to the right. Look to the right. Look to the right. Look to the right. Perfect. Now back onto the gas. We need to get up this hill. So let's add a little bit. Stay middle of the track. Middle. Middle. Middle. Now start steering. Now watch. The car is going to push out. Let it push out. Let it push out. And then full throttle. Okay? Fast hands here. Lift off the gas. Quick. Quick. Quick. That's okay. So stay to the right. This is the sector for you. go left. Aim for the tower. Aim for the tower. Perfect. Stay to the left. Stay to the left. Now to the right. Add gas. Perfect. Already look at the next cone. Look beyond it. Look for the turn. Look for the next cone. Look for the next one. Perfect. Stay to the right hand side. Add some gas. Now brake. Brake. There you go. Now add gas. Over to the left hand side. Aim for the tower. Aim for the tower. Aim for the tower. Now start going over to the cone. There you go. You got it. You got it. Perfect. Now a nice brake here. Keep looking right. Keep looking right. Keep looking right. Look beyond. perfect Stay to the right hand side. 
    Wait for the turn. Now turn. There you go. And then full. The track is divided into 10 segments (each corresponding to a corner). Here are the segment completion times: Segment 1: 14.13s, Segment 2: 11.83s, Segment 3: 17.36s, Segment 4: 7.90s, Segment 5: 12.84s, Segment 6: 10.42s, Segment 7: 9.15s, Segment 8: 7.69s, Segment 9: 7.60s, Segment 10: 6.82s. The following are the smoothness scores for each segment: Segment 1: 8.42, Segment 2: 14.19, Segment 3: 23.06, Segment 4: 10.35, Segment 5: 7.46, Segment 6: 10.22, Segment 7: 12.78, Segment 8: 13.00, Segment 9: 11.84, Segment 10: 27.06."
    }

\vspace{0.25cm}
    \end{minipage} 
    \hfill
    \begin{minipage}{0.49\textwidth}
    \tiny
    
    LLM output:
    \texttt{
    "Great job focusing on looking ahead and anticipating the turns. Your ability to predict the car's movement based on your sight is improving, especially in the middle sections of the track. Segment 3 was a bit rough, so let's work on maintaining smoother inputs there. Keep practicing those fast hands and smooth transitions between throttle and brake. Remember, the smoother you are, the more speed you can carry through the corners. Aim to reduce the steering inputs and keep the car balanced. Overall, you're doing well, just keep refining those techniques and you'll see even better lap times. Nice work!"
    }
    \end{minipage}    
    \hfill
    
    \caption{Prompt containing concurrent feedback text and additional lap level information and the terminal feedback response generated by the LLM. With additional information, the feedback is more specific and grounded.}
    \label{fig:icl_bsd}
\end{figure}

\subsection{Concurrent feedback imitation learning}
AI-assisted tutors are being developed for physical domains such as handwriting, driving and sports such as table tennis. A simple approach adopted in \cite{gopinath2024computational} and  ~\cite{ma2025t3set} is to frame this as learning a computational teaching policy. 

\textbf{Problem Statement:} Teaching policy can be modeled as a map between local context and behavior and the emitted teaching action. For race driving, the context is the local track map, represented as polylines of the left and right track boundaries, and driving trajectory, represented as a sequence of the vehicle positions over the last few seconds~\cite{gopinath2024computational}. We assess our dataset's utility for imitating coach instructions delivered promptly after the driving sequence. We further gauge the dataset utility in predicting the student driver's future trajectory, represented as a sequence of vehicle positions over a finite future horizon, following the same context. 

\textbf{Model Structure}: We utilize a standard encoder-decoder architecture with a shared latent representation for all decoder heads (coach instruction imitation and trajectory prediction). 

\begin{figure}[H]
 \begin{minipage}{\textwidth}
        \centering
        \includegraphics[width=0.95\linewidth]{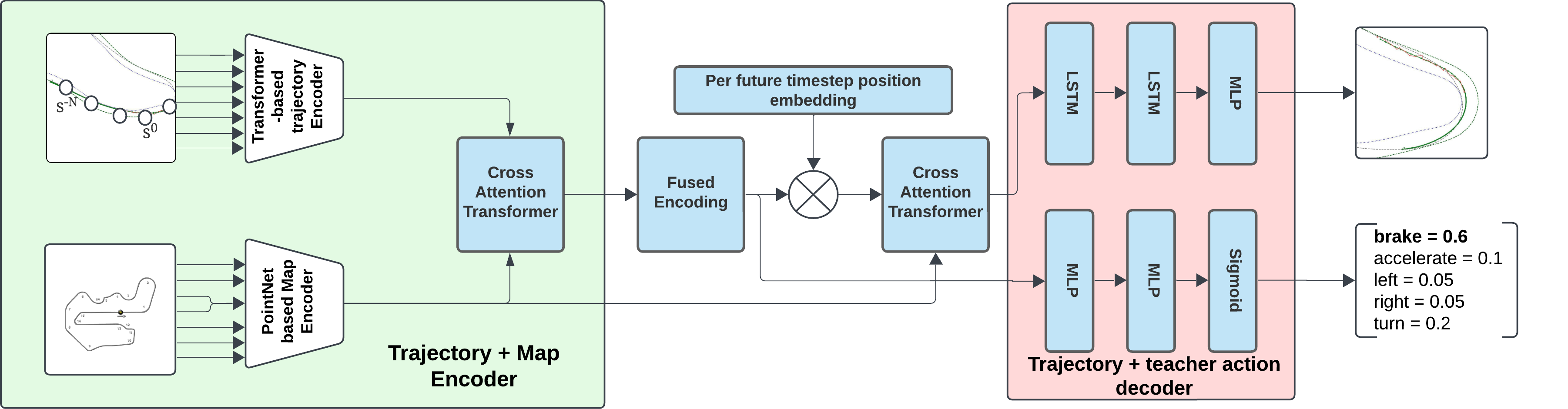}
    \end{minipage}
    \caption{Model architecture for computational teaching and multi-task learning. The inputs consist of a sequence
of past trajectories, encoded via a trajectory encoder consisting of an MLP+Transformer block, and the corresponding local map (along with racing line information) representations encoded via a PointNet based map encoder. The trajectory and map encodings are fused via a cross-attention mechanism to generated an encoded state. Per future timestep encoding is integrated into the fused encodings to decode the future trajectories, via an LSTM based decoder. The teacher actions are decoded from the fused trajectory and map encoding \textit{before} integration with per time-step position embeddings. as a multilabel target using a MLP based decoder with a sigmoid activation as the last layer.}
   \label{fig:model_diagram}
\end{figure}

Our baseline model consists of four primary components:
\begin{enumerate}
    \item \textbf{Trajectory encoder:} A multilayer perceptron (MLP) projects the input past trajectory into a latent space, followed by two Transformer blocks that encode the temporal dimension.
    \item \textbf{Map encoder:} A PointNet-based encoder processes the map features. Transformer layers are added after the pooling operation to capture the sequential structure of the track boundary lines.
    \item \textbf{Trajectory decoder:} An LSTM-based decoder is used, along with additional Transformer decoder layers to incorporate past trajectory and map information. 
    \item \textbf{Instruction Classifier:} A simple MLP-based decoder with sigmoid activations at the final layer does a multilabel classification task (over $N=12$ teaching categories), predicting the presence or absence of a coaching category in the future time-horizon (5s).
\end{enumerate}
All neural components use a hidden dimension size of 32. The encodings produced by the trajectory and map encoders are fused via a cross-attention mechanism. The fused encoding is then combined with per-timestep future position embeddings and cross-attended with the map encodings and decoded into a predicted trajectory. The teacher action decoding operates directly on the fused trajectory and map encoding \textit{before} integration with per time-step position embeddings. 
This architecture allows to jointly encode dynamic agent trajectories and static map features, while also providing instruction-level classification signals in a supervised manner.

\subsubsection{Input features}
The model ingests the past trajectory (5s) and a local map as input sampled at 5Hz. Trajectory coordinates are normalized relative to the first timestep of the past trajectory, and the same transformation is applied to the local map. Additional inputs include vehicle velocity, yaw, brake, throttle, steering, and the cone positions on the track. Note that cones are placed around the track during a high performance training session as visual markers to guide the gaze behavior of the student.

\begin{table}
  \caption{The full instruction set used for the multilabel teacher action prediction.}
  \label{tbl:instruction_categories}
  \centering
  \begin{tabular}{c|c}
    \toprule
                          Category Name      &    Description\\

    \midrule
     stay/move right     &  move to or stay on right side of the track            \\
     stay/move left     &  move to or stay on left side of the track        \\
     throttle-on & step on the throttle to accelerate \\
     throttle-off & foot off the throttle to slow down \\
     throttle-stay & maintain the throttle position \\
     brake-on & apply brake to slow down \\
     brake-off & step off brakes after application \\
     turn-left & turn the vehicle to the left \\
     turn-right & turn the vehicle to the right \\ 
     steering-landmark & steer vehicle towards a landmark (cone) \\
     steering-less & reduce the amount of steering input \\
     steering-little & apply small amounts of steering input \\
    \bottomrule
  \end{tabular}
\end{table}

\subsubsection{Model outputs}
\begin{figure}[H]
 \begin{minipage}{\textwidth}
        \centering
        \includegraphics[width=0.45\linewidth]{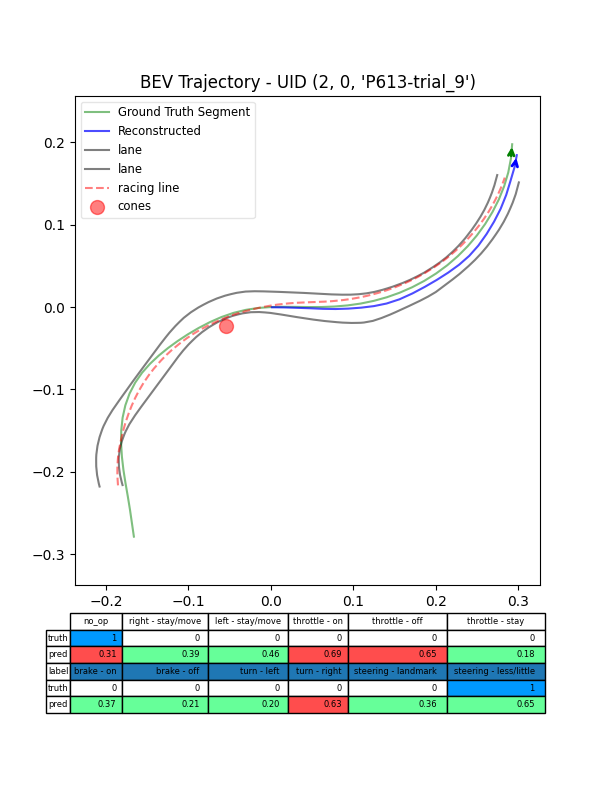}
    \end{minipage}    
   \vspace{-0.8cm}
    \caption{Example student trajectory prediction with multilabel prediction on teacher instruction categories. We observe that the predicted trajectory lines up quite well with the ground truth. The teacher action prediction contains a few false positives which are still contextually relevant for the future time horizon.}
   \label{fig:teacher_imitation}
\end{figure}

For trajectory prediction, the model outputs the future trajectory (5s) in the 2D plane. We cast the problem of teacher action prediction as a \textit{multilabel} classification problem over $N=12$ (compared to the six categories in Section\ref{statistics:cf},  we use a more fine-grained categorization) instruction categories, in which the label represents the presence or absence of a teaching category in the future time horizon. The complete set of instruction categories is given in Table~\ref{tbl:instruction_categories}. In the example prediction in Figure \ref{fig:teacher_imitation} we observe that the predicted trajectory tracks ground truth quite well and the predicted teacher actions are contextually relevant. See Appendix~\ref{appendix:prediction_examples} for additional examples.
\subsubsection{Training Details}
The imitation learning model is trained using the AdamW optimizer with an initial learning rate of $1 \times 10^{-3}$ and a linear learning rate decay schedule over 300 epochs, totaling 56,000 global steps with a batch size of 256. A dropout of 0.1 is applied. All transformer blocks use four attention heads and three transformer layers. Training is conducted on an AWS SageMaker \texttt{ml.g6.xlarge} instance.

\subsubsection{Loss Functions}: We use a MSE loss for trajectory prediction (weight coefficient = 320.0) and a weighted Binary Cross Entropy (wBCE) loss for teacher action prediction (weight coefficient = 1.0). Both weights are non-zero only in the multi-task training setup. The loss weights for the individual classes used in the wBCE loss are computed from the training dataset label statistics.

\subsubsection{Evaluation Criteria}
We evaluate model prediction by computing weighted $F_1$-score on the multilabel teacher action prediction and Root Mean Square Displacement for the auxiliary task of trajectory prediction. Due to the small size of the dataset, we rely on 15-fold cross validation with 5 random seeds on each fold and report metrics averaged across these runs. 

Table~\ref{tab_teaching_il} presents the results on both coach instructions' behavior cloning,  and students' trajectory prediction. We compare the performance of multi-task and single task training as in~\cite{gopinath2024computational}. We observe that under the multi-task setting, teacher action prediction slightly benefits from the trajectory prediction task. As a baseline, we ran a naive nearest neighbor baseline in which the teacher action associated with the average of 10-nearest neighbors was used as the prediction for each validation set data sample resulting in a weighted $F_1$-score of $0.498 \pm 0.030$ indicating that the neural approach gives us an almost $10\%$ improvement. 

\begin{table}
  \caption{Coach imitation and student trajectory prediction (single and multi-task training) from 15-fold cross-validation with 4 seeds each, along with standard deviations.}
  \label{tab_teaching_il}
  \centering
  \begin{tabular}{l|cc}
    \toprule
                           &   Coach imitation         &  Trajectory Prediction  \\
                           &  weighted $F_1$-score $\uparrow$       & RMSE (m) $\downarrow$  \\
    \midrule
    single-task (baseline)     & 0.584 $\pm$ 0.038       &    \textbf{0.249 $\pm$ 0.049}          \\
    multi-task     & \textbf{0.593 $\pm$ 0.038}	      & 0.250 $\pm$ 0.052           \\
    \bottomrule
  \end{tabular}%
\end{table}

\section{Limitations}

Our dataset has several limitations to acknowledge. First, our data is instruction in a racing simulator rather than a physical vehicle. Another aspect that is limited is the observability of the student -- while the coach was allowed to see the student, our dataset contains the vehicle state, transcripts, and additional annotations and metadata, rather than the full video of the interaction, which would have had privacy implications. Finally, the scope of the data is limited to a single coach teaching a small number of students ($n$=29) on the same track, making it an expert-behavior dataset from the coach side, not a population-level exploration.

\section{Conclusions}
We introduced \name{}, a first-of-its-kind dataset with longitudinal examples of embodied, motor learning with coaching interactions. The dataset enables a variety of analyses and ML training opportunities, towards AI approaches that automate and assist coaching in performance driving specifically, and embodied motor learning domains more broadly. We hope to see this approach generalized towards exploring the effectiveness of coaching styles or additional motor skills in future work.

\begin{acks}
We thank Coach Jon Gomes for providing coaching to our participants.
\end{acks}

\bibliographystyle{plain}
\bibliography{sample-base}

\appendix

\section{Additional imitation learning examples}\label{appendix:prediction_examples}
\begin{figure*}
    \centering
    \begin{minipage}{0.3\textwidth}
        \centering
        \includegraphics[width=\linewidth]{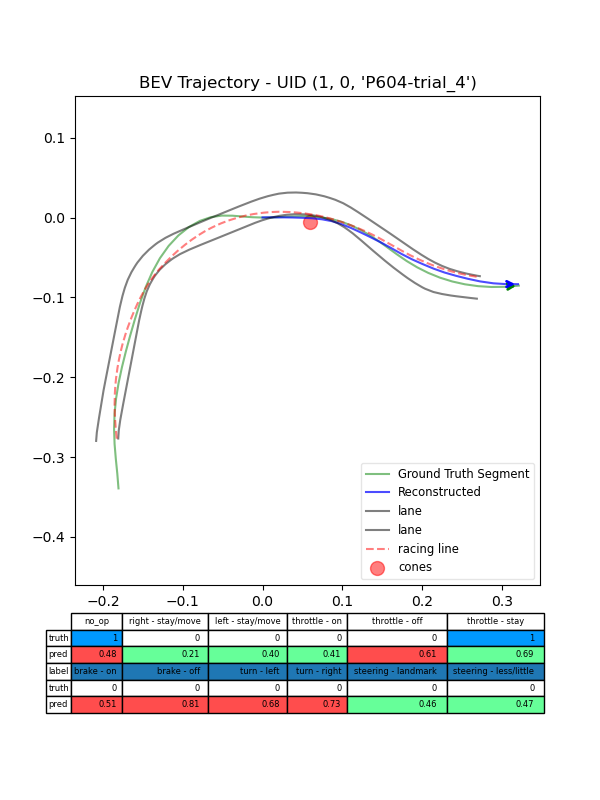}
        (a)
    \end{minipage}\hfill
    \begin{minipage}{0.3\textwidth}
        \centering
        \includegraphics[width=\linewidth]{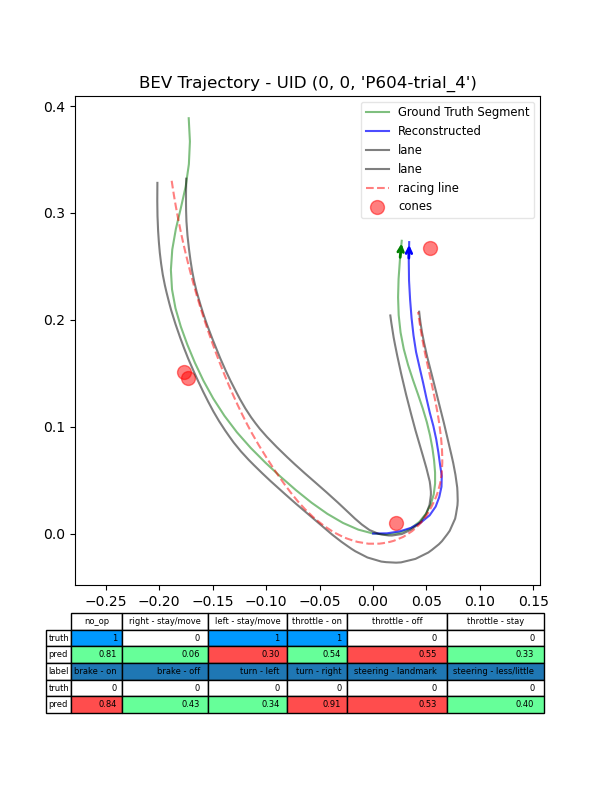}
        (b)
    \end{minipage}\hfill
    \begin{minipage}{0.3\textwidth}
        \centering
        \includegraphics[width=\linewidth]{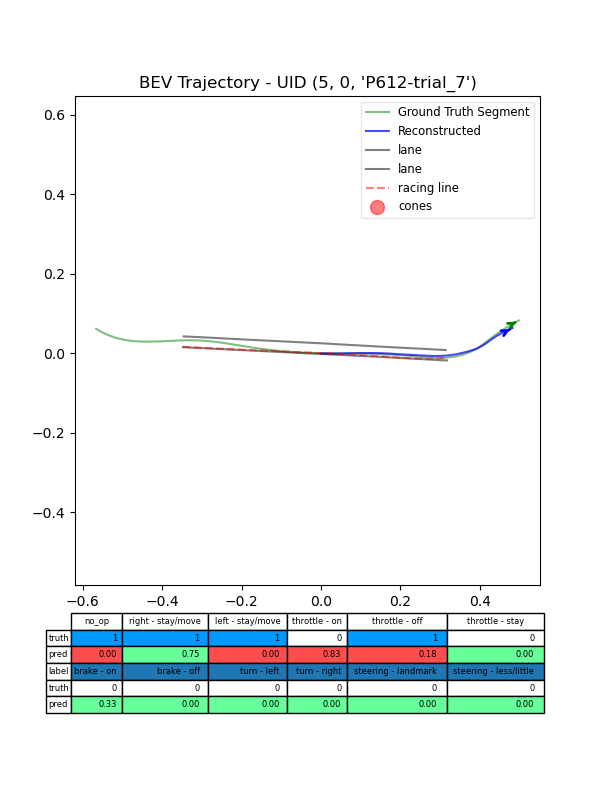}
        (c)
    \end{minipage}
    \caption{Additional examples of student trajectory prediction, and teacher action multilabel predictions (over twelve categories).}
    \label{fig:teacher_imitation_appendix}
\end{figure*}
In this subsection, we provide additional examples of trajectory prediction and teacher action prediction in Figure~\ref{fig:teacher_imitation_appendix}. In these examples, we can see how the model has learned to correlate trajectory and map features to perform quite well on trajectory prediction (as qualitatively seen by match between predicted trajectory and the ground truth trajectory. Despite false positives in the teacher action prediction, we still see that some of the false positives such as (`throttle - off' and `throttle - on') when entering an exiting a corner complex are still meaningful and contextually relevant in the high performance driving, for example in (Figure~\ref{fig:teacher_imitation_appendix}(a))

\section{Additional in-context learning details}
\label{appendix:icl_prompts}
For in-context learning experiments, we used the following base task prompt. 

\begin{lstlisting}
You will be given counts of instruction categories reflecting instructions given to a novice student by a driving coach during a high performance driving lap.
The categories are: 
    1. Throttle refers to the student engaging the throttle/gas or talking about speed
    2. Lateral position refers to the positioning of the car on the track 
    3. Turn refers to car aiming especially around corners
    4. Steering refers to steering wheel inputs
    5. Brake refers to activating the brake to slow or speed or control the car
    6. Looking ahead refers to directing their sight ahead of the immediate track to look for turns or cones
    7. Other refers to any other instruction not included in the other 6 categories.
 Your task is to predict the coach's terminal feedback after the lap. Use clues in the instruction category counts and from the context information (if present) to inform the content of the terminal feedback. Note how not all categories are always explicitly mentioned.         
Present responses in a structured in JSON format with a single key called 'feedback'. Please do not have any nested structures in your response. Use language typically used for high-performance driving education. Keep it as specific as possible
\end{lstlisting}
The following is an example of GPT-generated terminal feedback when no additional information is and no in-context examples are provided. The task prompt is omitted for brevity.
\begin{lstlisting}
GPT Prompt: 
"P603's trial 11 concurrent feedback contained the following: Little turn to the left. Start lifting off the gas. Look to the right. Look to the right. Look to the right. Look to the right. Perfect. Now back onto the gas. We need to get up this hill. So let's add a little bit. Stay middle of the track. Middle. Middle. Middle. Now start steering. Now watch. The car is going to push out. Let it push out. Let it push out. And then full throttle. Okay? Fast hands here. Lift off the gas. Quick. Quick. Quick. That's okay. So stay to the right. This is the sector for you. go left. Aim for the tower. Aim for the tower. Perfect. Stay to the left. Stay to the left. Now to the right. Add gas. Perfect. Already look at the next cone. Look beyond it. Look for the turn. Look for the next cone. Look for the next one. Perfect. Stay to the right hand side. Add some gas. Now brake. Brake. There you go. Now add gas. Over to the left hand side. Aim for the tower. Aim for the tower. Aim for the tower. Now start going over to the cone. There you go. You got it. You got it. Perfect. Now a nice brake here. Keep looking right. Keep looking right. Keep looking right. Look beyond. perfect Stay to the right hand side. Wait for the turn. Now turn. There you go. And then full.

Response:
"Great job on maintaining focus and looking ahead, which is crucial for anticipating track changes. Your throttle control was smooth, especially when adding gas on the straights and lifting off when needed. Keep practicing your lateral positioning; staying middle and then transitioning to the sides was well executed. Steering inputs were responsive, but continue to work on fast hands for quicker adjustments. Braking was effective, especially in the sectors where you needed to slow down before turns. Overall, your awareness and anticipation of track elements are improving. Keep refining your technique and aim for smoother transitions between throttle, brake, and steering inputs."

\end{lstlisting}

In this example, we see that the terminal feedback is quite generic and touches upon on all aspects of high-performance driving. As there is no additional information, the response is not specific, however it reflects the main driving themes that the coach emphasized during the lap. In particular, is the model's commentary on anticipation of track elements in light of the `looking ahead' related instruction provided during the lap. 

We also experimented with Claude Opus 4.7 via Amazon Bedrock API. We observed that despite providing segment level metrics such as segment completion times and smoothness scores, Claude's responses were not grounded in segment specific information. However, the model is able to latch on to information present in the concurrent feedback. Claude API also offers the ability to invoke tool-use for structured responses. The following is the response from Claude for the same prompt used in GPT based example above (except with segment-level metrics). 

\begin{lstlisting}

Claude Opus 4.7 Response:
"Alright, so that one you did really well. You're starting to look ahead a lot more, which is huge. That's why you're able to react faster on the steering and the throttle. The braking zones, you're getting more comfortable with them. The only thing I want you to work on now is being a little smoother on the exit of those last couple of turns. You're getting back to the throttle, which is great, but just make sure your hands are nice and smooth as you unwind the wheel. Okay? Ready for the next one?"

\end{lstlisting}

\section{Additional instruction category distribution by location}\label{appendix:additional_instruction_category}

\begin{figure}[H]
    \begin{minipage}{0.45\textwidth}
        \centering
        \includegraphics[width=0.9\linewidth]{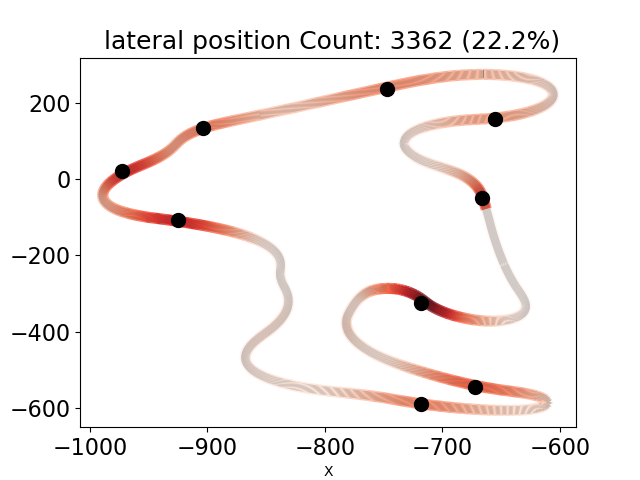}

        (a)
    \end{minipage}    
        \begin{minipage}{0.45\textwidth}
        \centering
        \includegraphics[width=0.9\linewidth]{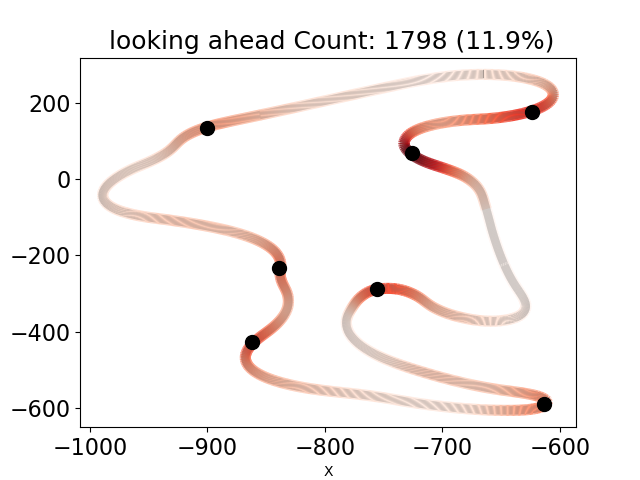}
    
        (b)
    \end{minipage}     
 \begin{minipage}{0.45\textwidth}
        \centering
        \includegraphics[width=0.9\linewidth]{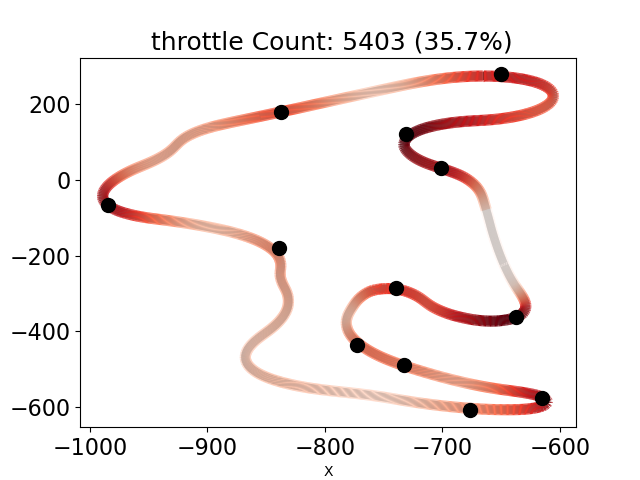}

        (c)
    \end{minipage}    
        \begin{minipage}{0.45\textwidth}
        \centering
        \includegraphics[width=0.9\linewidth]{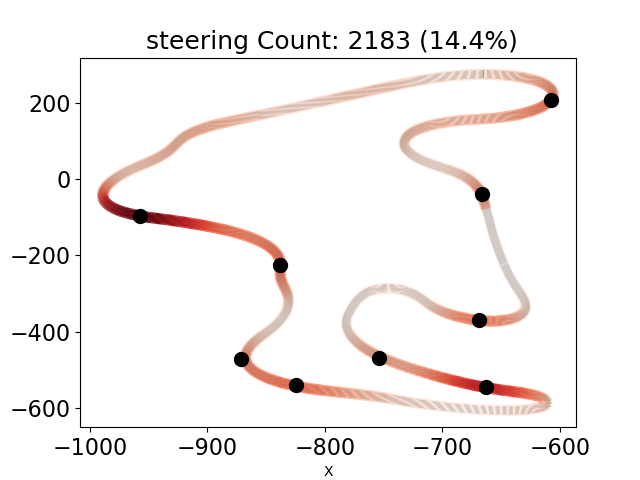}
    
        (d)
    \end{minipage}     
    \caption{Density of coach utterances along the track for the remaining four categories. The black dots in the figure are the discovered modes and exhibit high location specificity.}
    \label{fig:category-heatmap-appendix}
\end{figure}

Figure~\ref{fig:category-heatmap-appendix} shows the distribution of the remaining categories (lateral position, looking ahead, throttle, and steering) along the track. We observe that the modes of the different categories have high specificity and correlation to the geometrical features of the map. For example, the lateral placement of the car becomes important before approaching a corner as can be seen in Figure~\ref{fig:category-heatmap-appendix} (a). Similarly in Figure~\ref{fig:category-heatmap-appendix} (c), we observe that throttle related instructions are primarily relayed during the entry (throttle-off to slow down) and the exit (throttle-on to pick up speed again) phase of a corner. 
\section{Additional comparisons between coached vs. self-practice conditions}\label{appendix:additional_coached_self-practice}
\begin{figure}[H]
    \centering
    \begin{minipage}{0.45\textwidth}
        \centering
        \includegraphics[width=\linewidth]{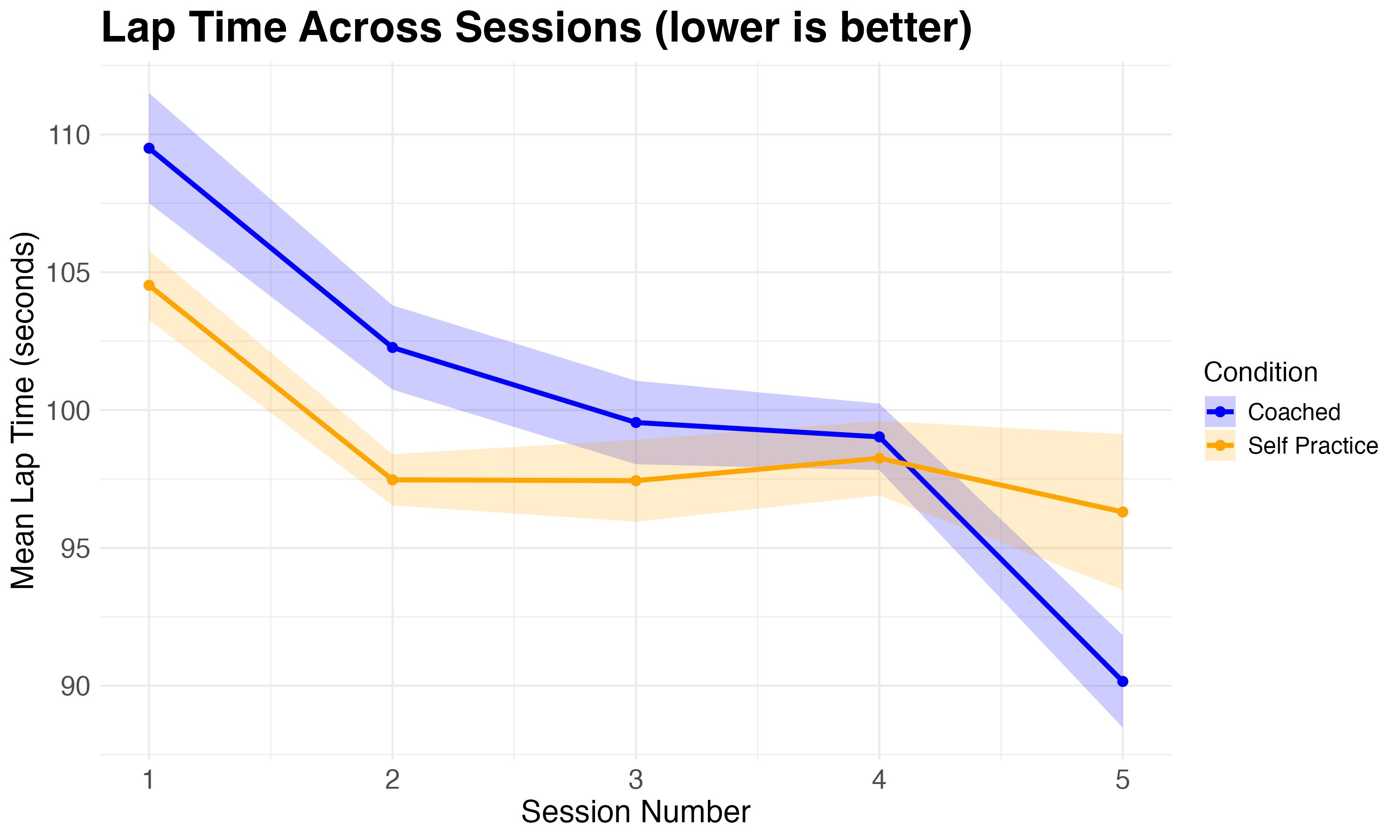}
        (a)
    \end{minipage}
    \hfill
    \begin{minipage}{0.45\textwidth}
        \centering
        \includegraphics[width=\linewidth]{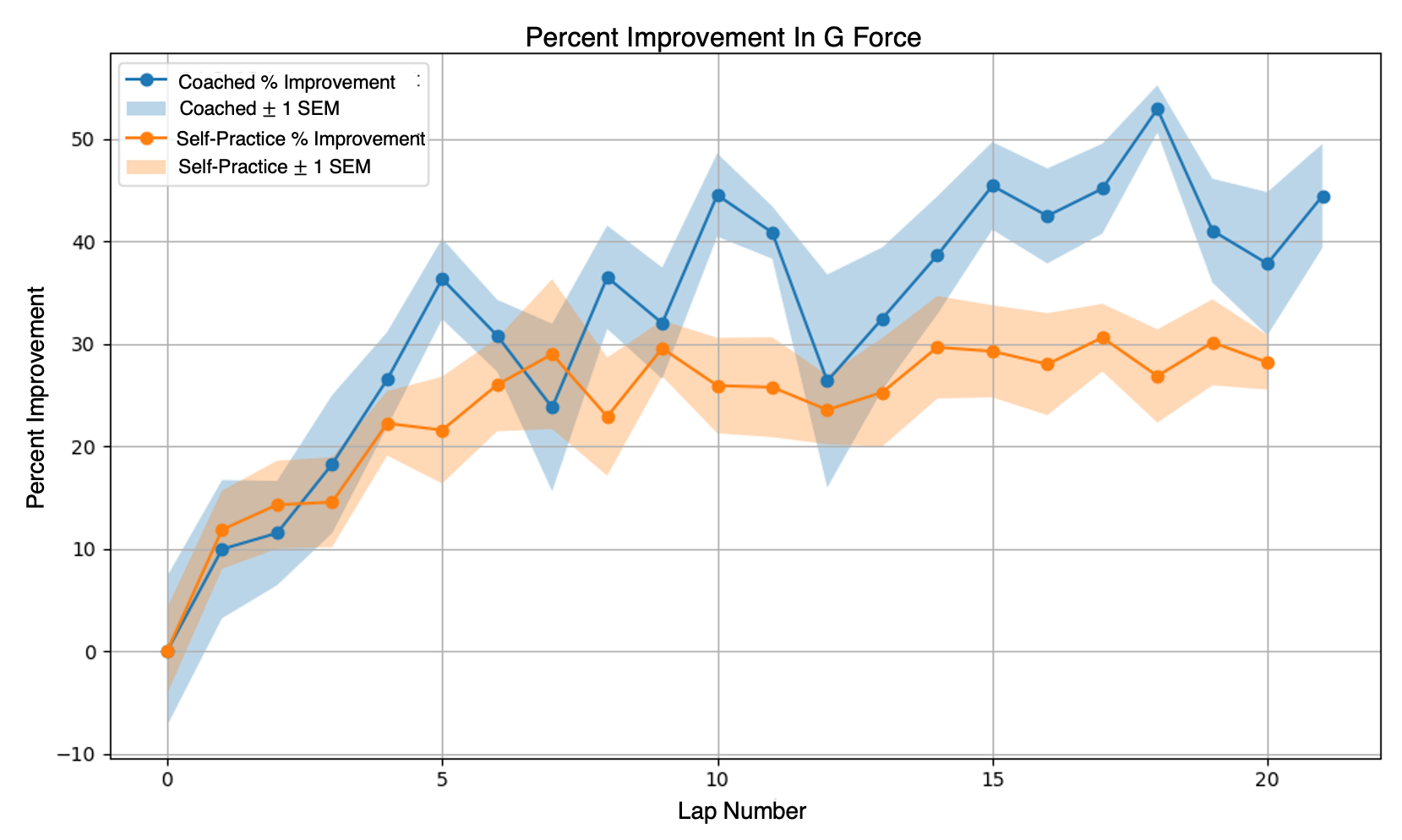}
        (b)
    \end{minipage}
    
    \vspace{1em} 
    
    \begin{minipage}{0.45\textwidth}
        \centering
        \includegraphics[width=\linewidth]{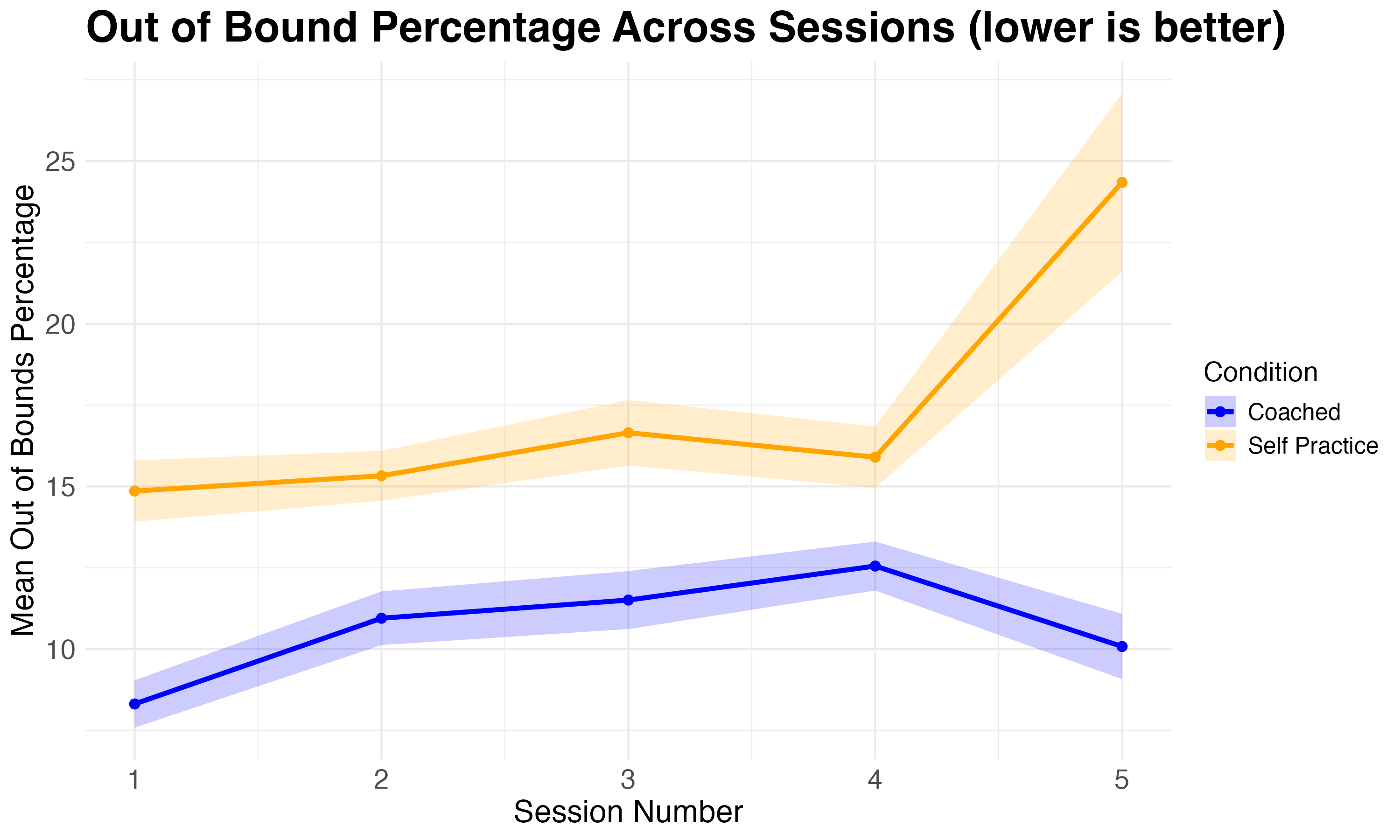}
        (c)
    \end{minipage}
    \hfill
    \begin{minipage}{0.45\textwidth}
        \centering
        \includegraphics[width=\linewidth]{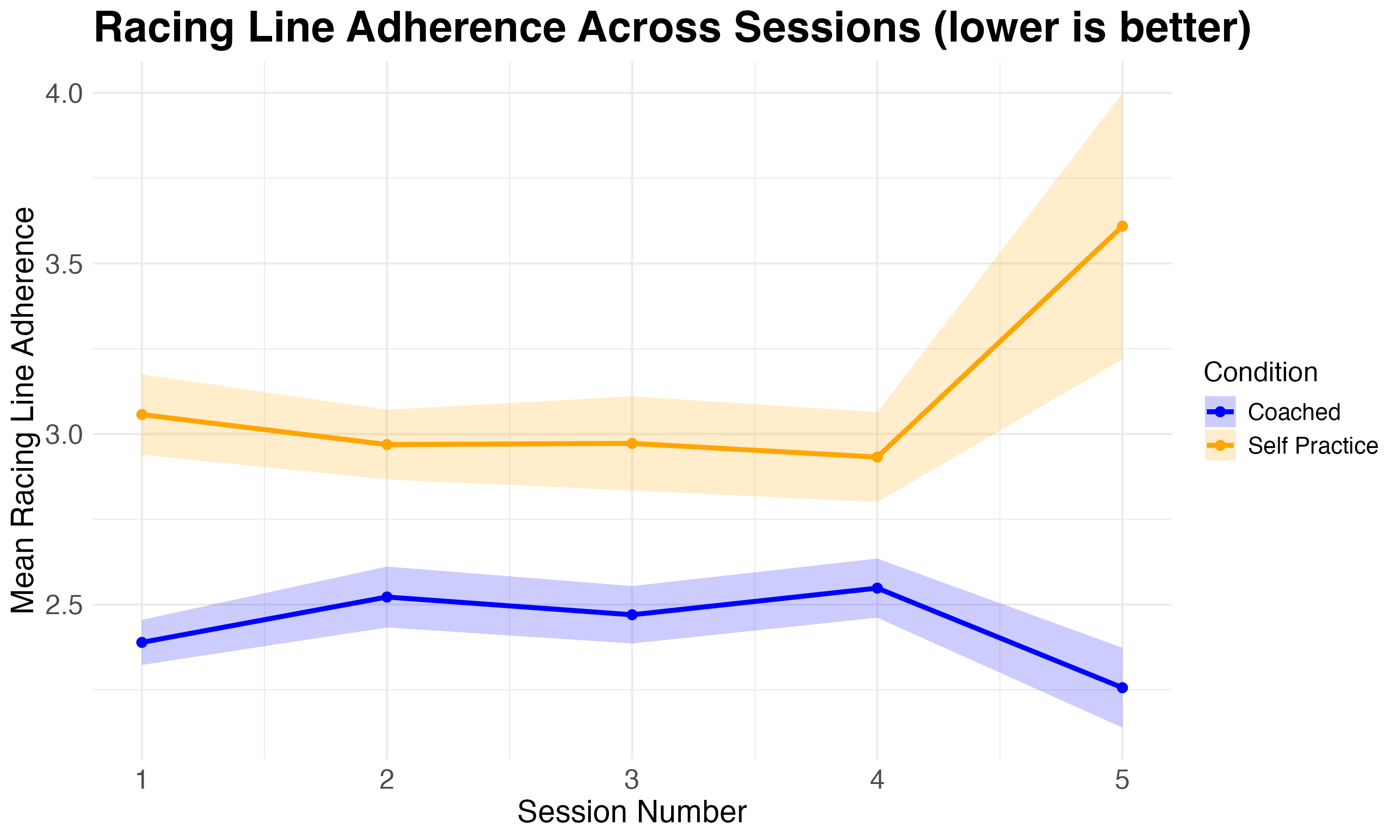}
        (d)
    \end{minipage}
    
    \caption{Self-practice vs coached students. (a) Lap time improvement over sessions (higher is better). (b) Percent improvement in g-force during cornering (higher is better). (c) Percentage of driving time with excursions out of the track bounds (lower is better). (d) Racing line adherence average distance (lower is better).}
    \label{fig:self-vs-coached}
\end{figure}

Figure~\ref{fig:self-vs-coached} shows (a) lap times across sessions, (b) percent G force improvement at turns, (c) race track excursions (car leaving the edge of the track typically due to lack of control or spinouts), and (d) racing line adherence over training time for both self-practice and coached participants. While the self-practice participants are typically faster than the coached participants, the coached participants spend less time off track and are able to stay closer to the raceline more often.  

\begin{figure}[H]
    \centering
    \begin{minipage}{0.45\textwidth}
        \centering
        \includegraphics[width=0.95\linewidth]{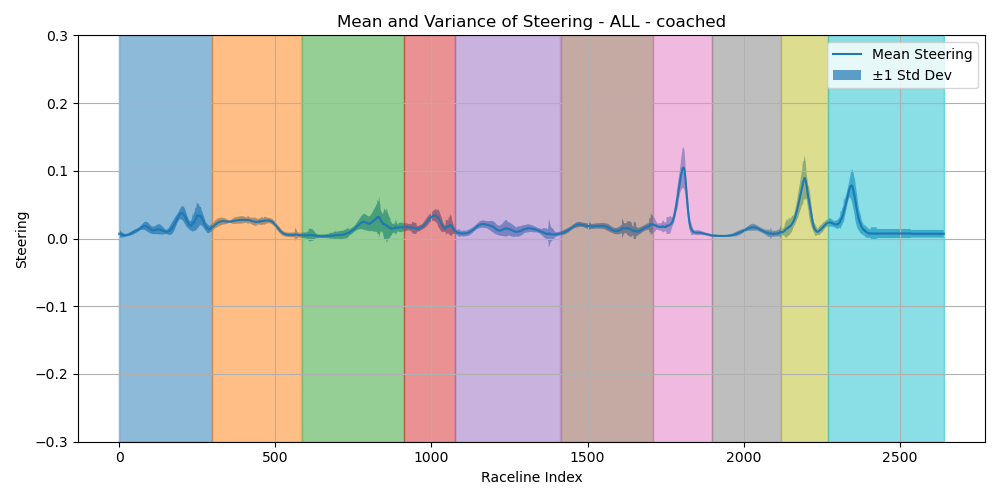}
        
        (a)
    \end{minipage}
    \hfill
    \begin{minipage}{0.45\textwidth}
        \centering
        \includegraphics[width=0.95\linewidth]{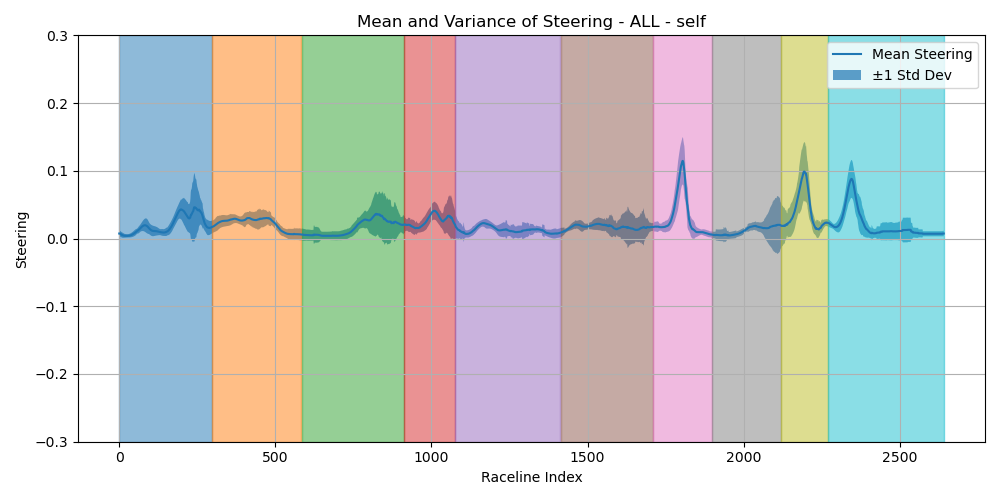}
        
        (b)
    \end{minipage}
    \hfill
    \begin{minipage}{0.45\textwidth}
        \centering
        \includegraphics[width=0.95\linewidth]{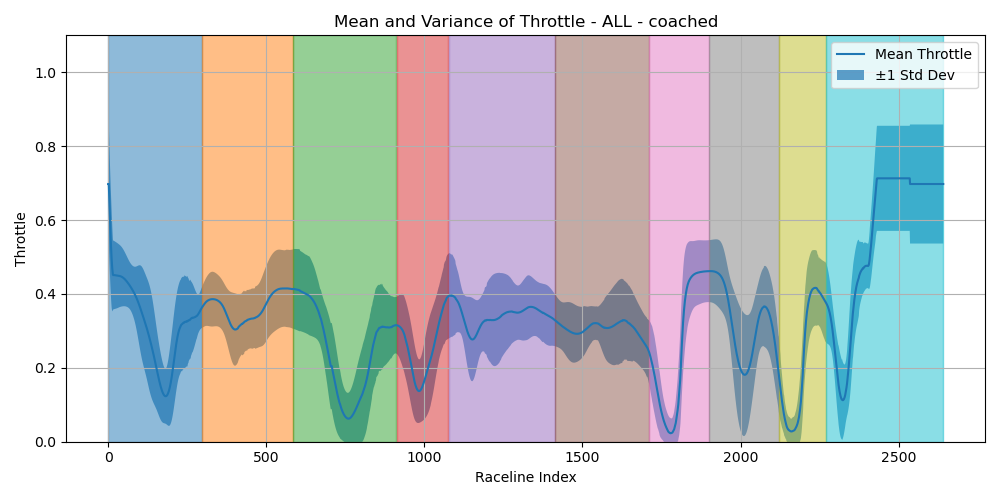}
        
        (c)
    \end{minipage}
    \hfill
    \begin{minipage}{0.45\textwidth}
        \centering
        \includegraphics[width=0.95\linewidth]{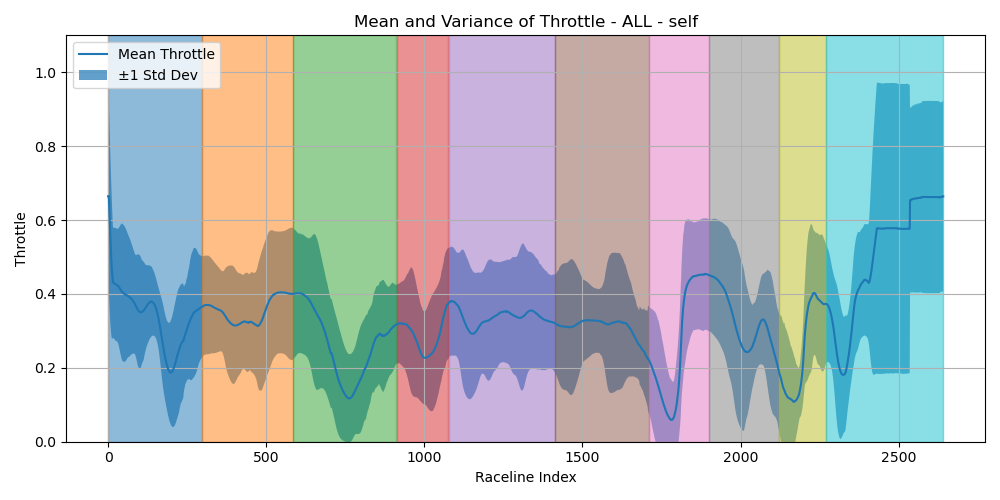}
        
        (d)
    \end{minipage}

    \caption{Self-practice vs coached students. (a) Mean and variance of steering profile across participants for the coached condition as a function of progress along the track. (b) Mean and variance of steering profile across participants for the self-practice condition as a function of progress along the track. (c) Mean and variance of throttle profile across participants for the coached condition as a function of progress along the track. (d) Mean and variance of throttle profile across participants for the self-practice condition as a function of progress along the track. Ribbons represent  +/- 1 standard error.}
    \label{fig:self-vs-coached-control}
\end{figure}

We also present the steering and throttle profiles as a function of progress along the track in for the coached group (Figure~\ref{fig:self-vs-coached-control}(a, b)) and the self-practice group (Figure~\ref{fig:self-vs-coached-control}(c, d)). Similar to the characteristics observed in the speed profile, we notice that the variance is higher for both steering and throttle indicating that in the self-practice group different subjects follow different strategies for navigating the track quickly. We also notice that the throttle profile for the self-practice group is noisier indicating less smooth throttle control. 
%
        
            
        
            

\section{Additional language visualizations}
\label{appendix:additional_language_statistics}
\begin{figure}[H]
 \begin{minipage}{0.5\textwidth}
        \centering
        \includegraphics[width=1\linewidth]{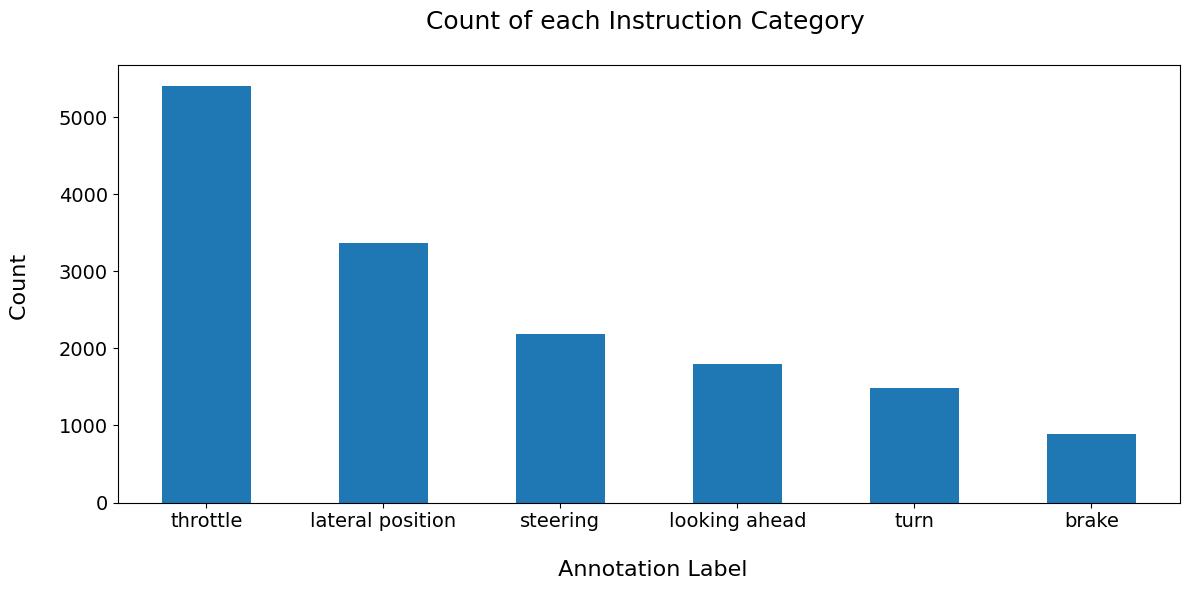}

        (a)
    \end{minipage}    
 \begin{minipage}{0.5\textwidth}
        \centering
        \includegraphics[width=1\linewidth]{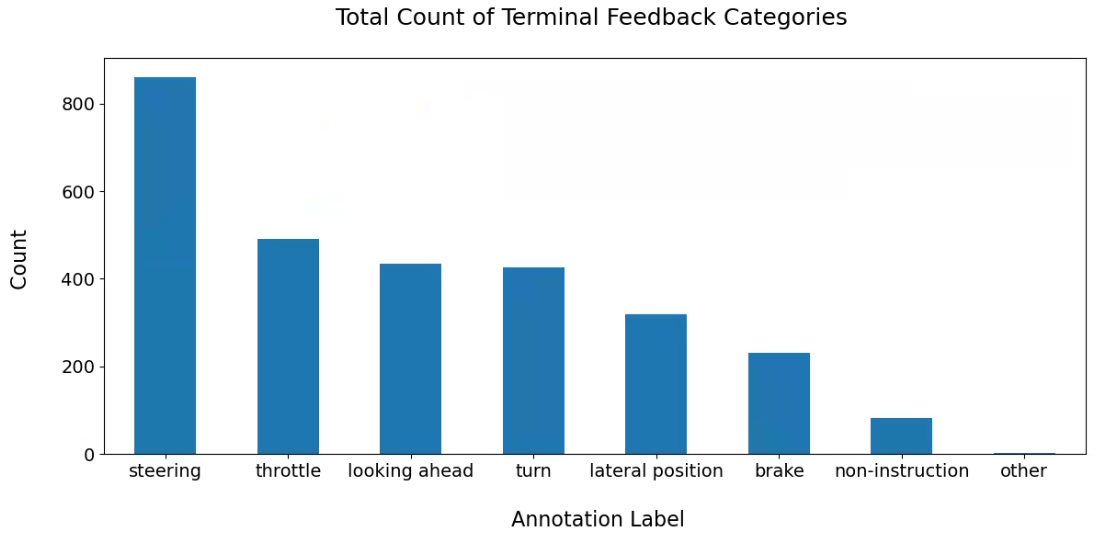}

        (b)
    \end{minipage}    

    \caption{(a) Concurrent instruction category counts. (b) Terminal feedback category counts. Notice the non-uniform distribution of different instructions.}
    \label{fig:concurrent-counts}
\end{figure}

Figure~\ref{fig:concurrent-counts} (a) shows the counts of different instruction categories in concurrent feedback, Figure~\ref{fig:concurrent-counts}(b) shows the counts of different instruction categories in the terminal feedback. While throttle is the most common instruction in the concurrent feedback, steering is the most common instruction discussed in the terminal feedback. Steering is also one of the more difficult categories to comply with, so that may be why it is discussed further (see Figure~\ref{fig:compliance-overview} in the main manuscript).

\begin{figure}[H]
        \begin{minipage}{0.45\textwidth}
        \centering
        \includegraphics[width=\linewidth]{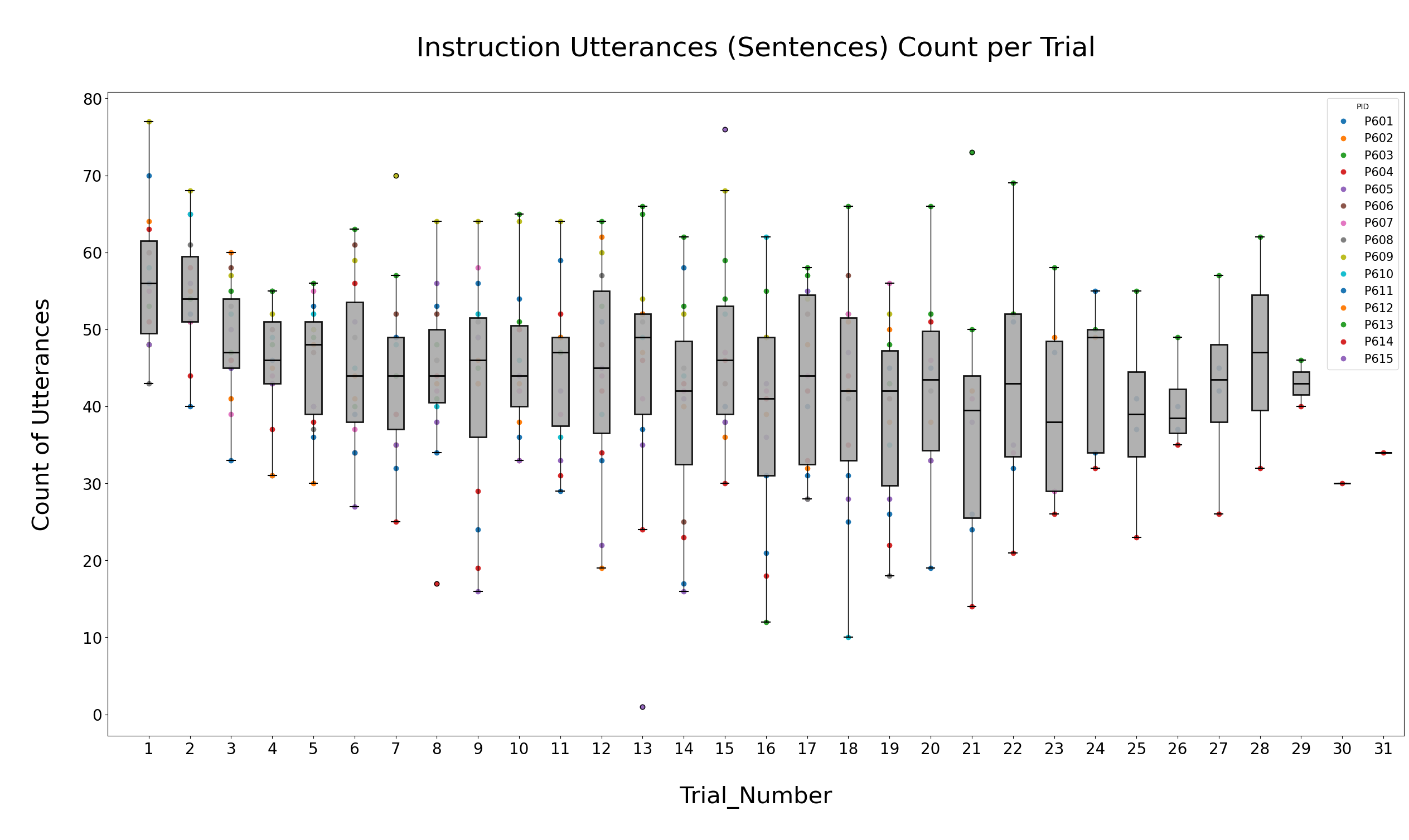}

    (a)
    \end{minipage}    
    \begin{minipage}{0.45\textwidth}
        \centering
        \includegraphics[width=\linewidth]{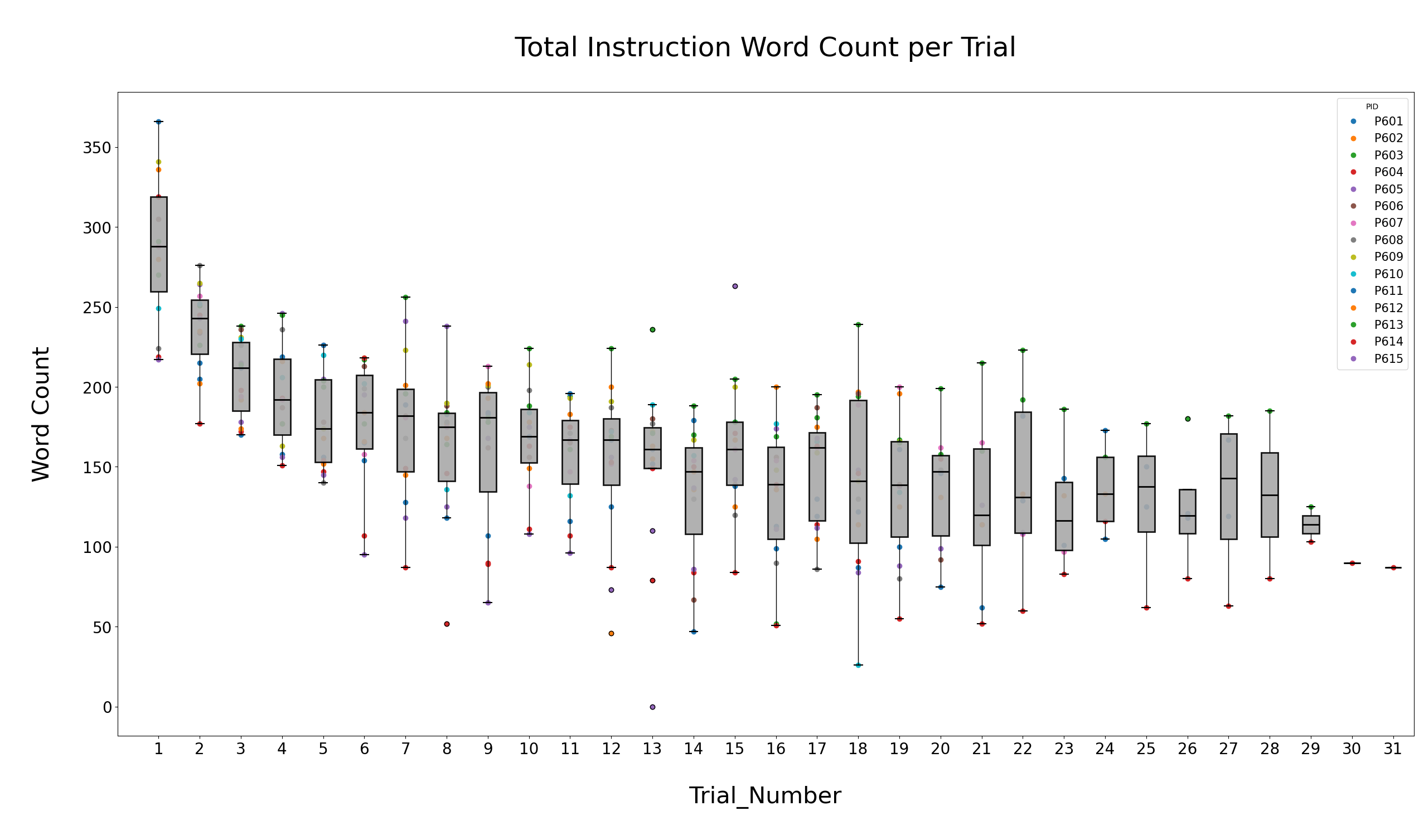}

    (b)
    \end{minipage}    

    \caption{(a) Instructions counts and (b) instruction word counts per trial. Note how both utterances and word counts decrease over time.}
    \label{fig:counts-per-trial}
\end{figure}

Figure~\ref{fig:counts-per-trial}(a) shows the distribution of average sentence counts per trial and Figure~\ref{fig:counts-per-trial}(b) shows the distribution of total word count per trial. There was a negative Pearson's correlation between both average sentence count and trial number ($r= -.64, p < .001.$) and word count per trial ($r= -.263, p < .001$).

\section{Additional topic modeling visualizations}

\begin{figure}[H]
    \centering
    \includegraphics[width=1\linewidth]{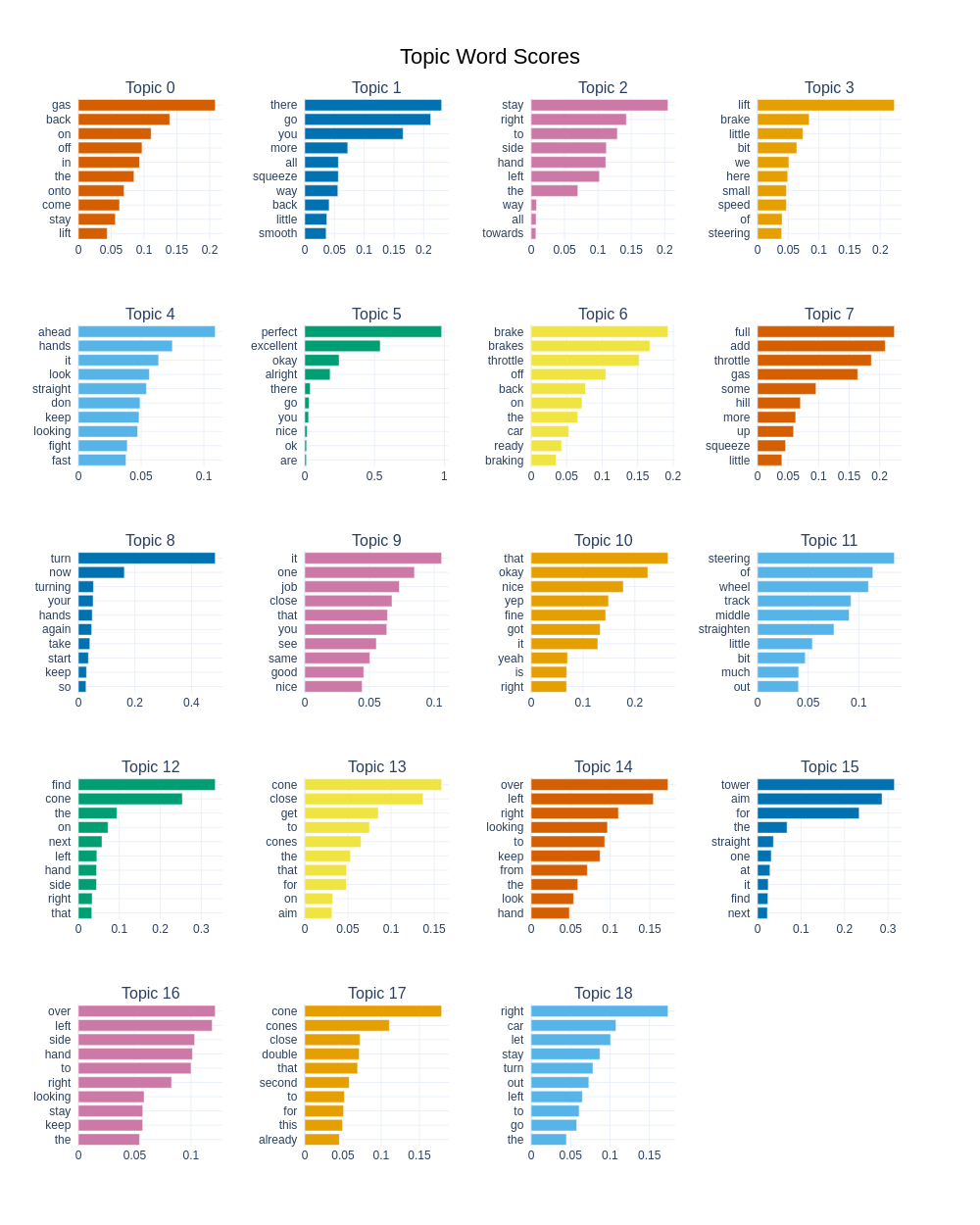}
    \caption{Word distribution in the topics discovered by BERTopic.}
    \label{fig:word_distribution_topic_appendix}
\end{figure}

Figure~\ref{fig:word_distribution_topic_appendix} indicates the word distribution in all of the topics discovered by BERTopic. We notice that HPDE related themes emerge in this topics and keywords associated with different high performance driving related skills (such as brake, throttle, looking ahead) are captured clearly in the topics. Additionally, BERTopic was able to discover topics related to positive feedback and encouragement as well.

\section{Additional student state statistics} \label{appendix:student-state-stats}

\begin{figure}[H]
    \centering
    \includegraphics[width=\linewidth]{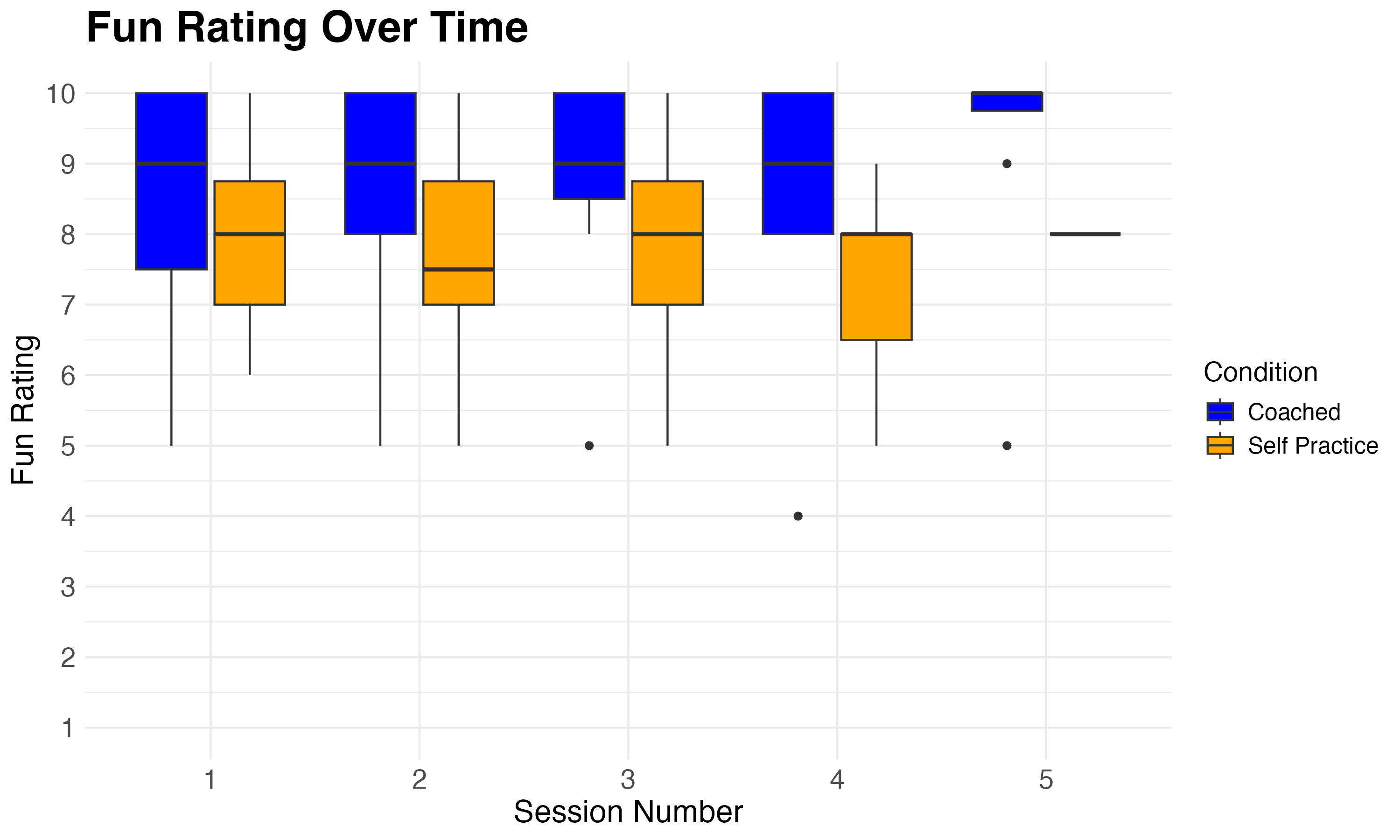}
    \caption{Box plots showing rating on the fun metric over time, broken down by condition.}
    \label{fig:fun-by-condtion-and-session}
\end{figure}

\begin{figure}[H]
    \centering
    \includegraphics[width=\linewidth]{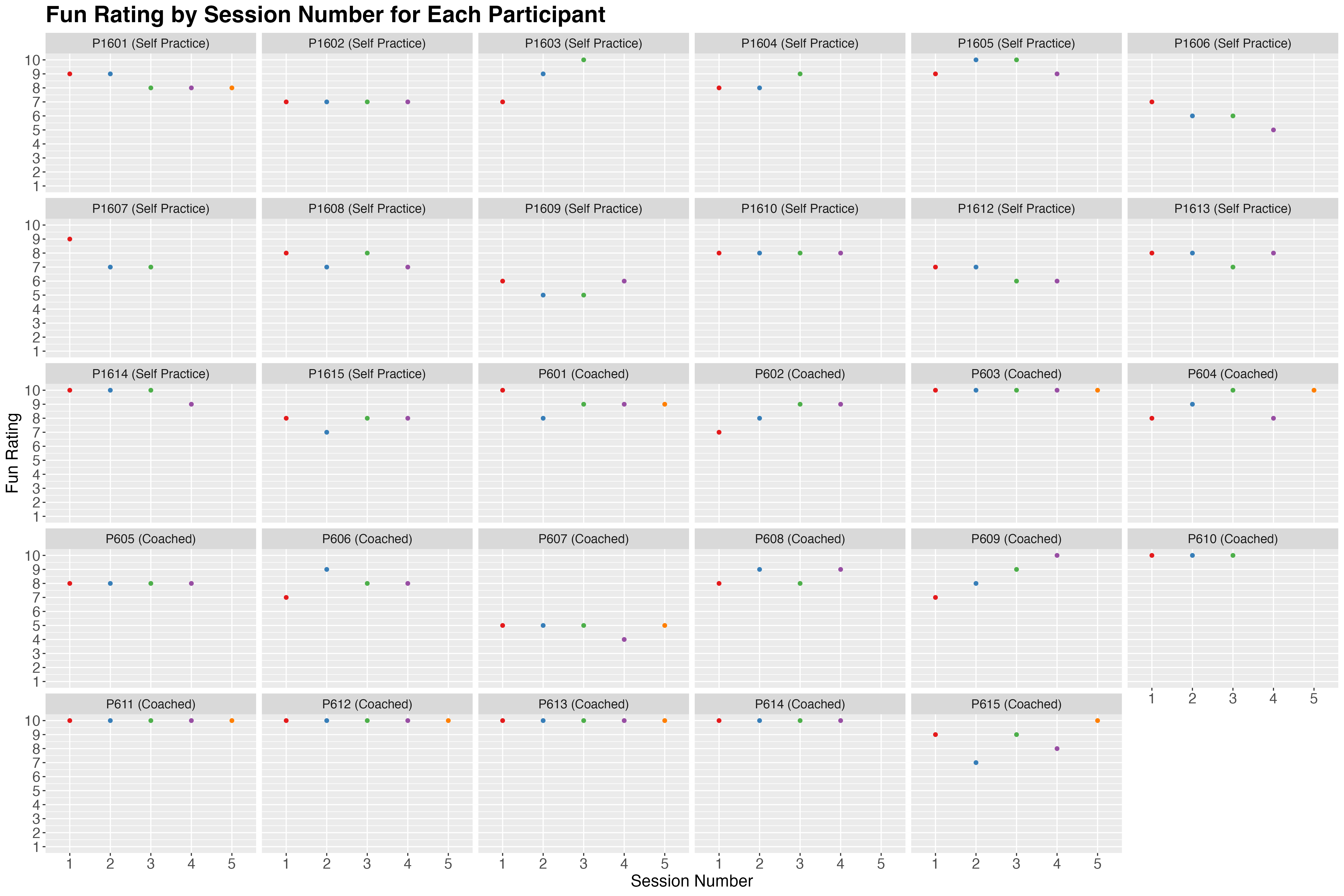}
    \caption{Individual plots looking at fun over time, broken down by participant.}
    \label{fig:fun-by-participant}
\end{figure}

Figure~\ref{fig:fun-by-condtion-and-session} and Figure~\ref{fig:fun-by-participant} show more details on the breakdown of fun by condition, time, and individuals.

\section{Additional compliance analysis statistics} \label{appendix:compliance-analysis-stats}
Tables~\ref{tab:compliance-lmm-fixed-effects} and \ref{tab:compliance-analysis-random-effects} show the linear mixed effects analysis for compliance as a factor of instruction type and trial number, suggesting that compliance increases over time, and that both lateral position and steering have lower compliance. Additionally, people improve less quickly over time on lateral position instructions.  

\begin{figure}[H]
    \centering
    \includegraphics[width=.5\linewidth]{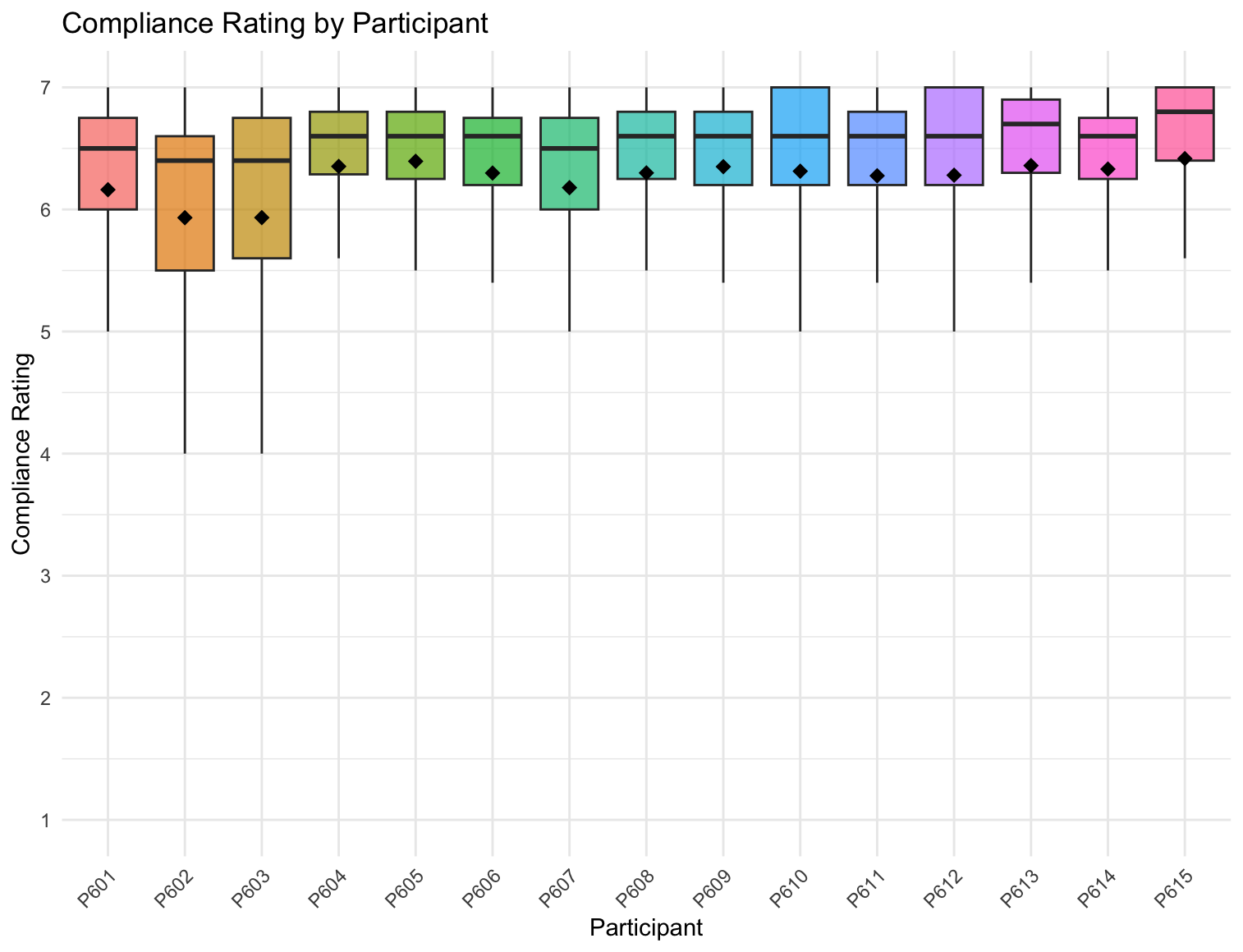}
    \caption{Individual differences across participants in compliance scores.}
    \label{fig:compliance-individual}
\end{figure}

\begin{figure}[H]
    \centering
    \includegraphics[width=.5\linewidth]{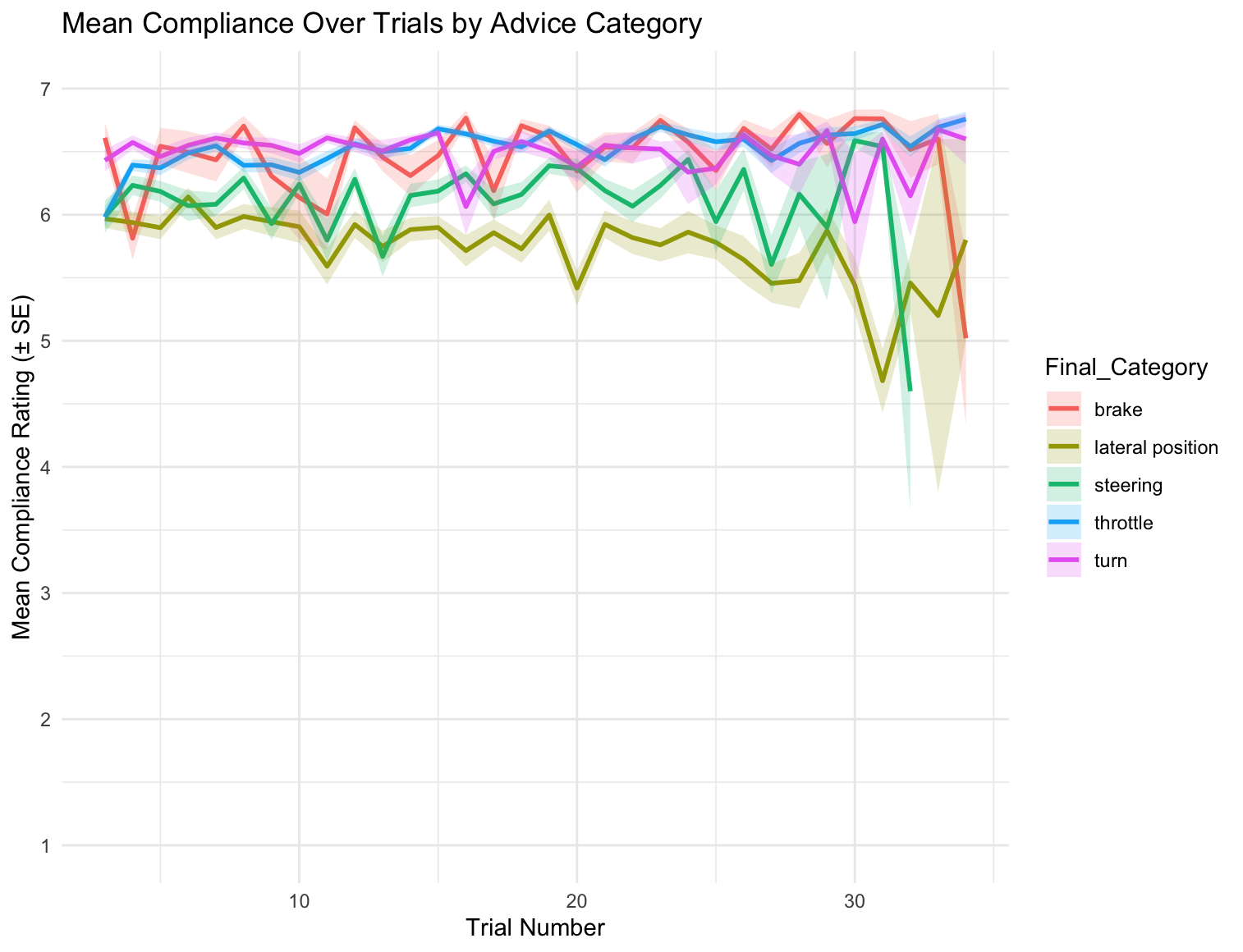}
    \caption{Differences in compliance scores across trials, ribbons represent +/-1 standard error.}
    \label{fig:compliance-trial}
\end{figure}

 Figure~\ref{fig:compliance-individual} shows the individual differences, and Figure~\ref{fig:compliance-trial} shows what each category looks like over time. 
\begin{table*}[ht]
\centering
\caption{Fixed Effects from Linear Mixed-Effects Model Predicting Compliance}
\label{tab:compliance-lmm-fixed-effects}
\begin{tabular}{lrrrrrl}
\toprule
\textbf{Predictor} & \textbf{Estimate} & \textbf{SE} & \textbf{df} & \textbf{$t$} & \textbf{$p$} & \textbf{Sig.} \\
\midrule
Intercept & 6.284 & 0.086 & 379.7 & 73.41 & $<$ .001 & *** \\
Trial Number & 0.011 & 0.005 & 10960 & 2.43 & .015 & * \\
Lateral Position & –0.283 & 0.085 & 10960 & –3.34 & $<$ .001 & *** \\
Steering & –0.250 & 0.091 & 10960 & –2.76 & .006 & ** \\
Throttle & –0.008 & 0.083 & 10960 & –0.09 & .927 &  \\
Turn & 0.210 & 0.096 & 10960 & 2.19 & .028 & * \\
Trial Number × Lateral Position & –0.023 & 0.005 & 10960 & –4.58 & $<$ .001 & *** \\
Trial Number × Steering & –0.006 & 0.005 & 10960 & –1.06 & .289 &  \\
Trial Number × Throttle & 0.004 & 0.005 & 10950 & 0.80 & .422 &  \\
Trial Number × Turn & –0.011 & 0.006 & 10950 & –2.00 & .046 & * \\
\bottomrule
\end{tabular}
\vspace{1ex}
\begin{flushleft}
\textit{Note}. SE = Standard Error. df = degrees of freedom. Sig. codes: *** $p < .001$, ** $p < .01$, * $p < .05$, . $p < .10$
\end{flushleft}
\end{table*}

\begin{table}[ht]
\centering
\caption{Random Effects from Linear Mixed-Effects Model}
\label{tab:compliance-analysis-random-effects}
\begin{tabular}{lrr}
\toprule
\textbf{Group} & \textbf{Variance} & \textbf{SD} \\
\midrule
Participant (Intercept) & 0.020 & 0.140 \\
Residual & 0.765 & 0.875 \\
\bottomrule
\end{tabular}
\end{table}

\section{Compliance annotation instructions} 
This section provides additional details on the annotation procedures followed for compliance annotations. 
\label{appendix:comp-annotation}

\subsection{Task description}
In this task, you will watch a short video clip of a coach helping a student to improve their performance-driving skills. You will answer questions about the coach’s instructions and the student’s response to the instructions. 
The objective of this annotation is for us to learn more about how effectively students are following the coach’s instructions, so please pay attention to the driver’s control (on the bottom of the screen) and how well they appear to be listening to the coach. Remember that you can rewind/ watch the video multiple times if you need to rewatch it.
For each video, you will hear the coach give several pieces of advice or feedback. Only consider the advice that is written underneath the video. We only want to have the first utterance labeled.
The first utterance will be the one that you hear being given just after the video starts (the videos are programmed to start 0.5 seconds before the coach starts speaking). So the first complete command that you hear should match what is printed under the video, and that is what we want to label.

\begin{figure}[H]
    \begin{minipage}{0.48\textwidth}
        \centering
        \includegraphics[width=0.9\linewidth]{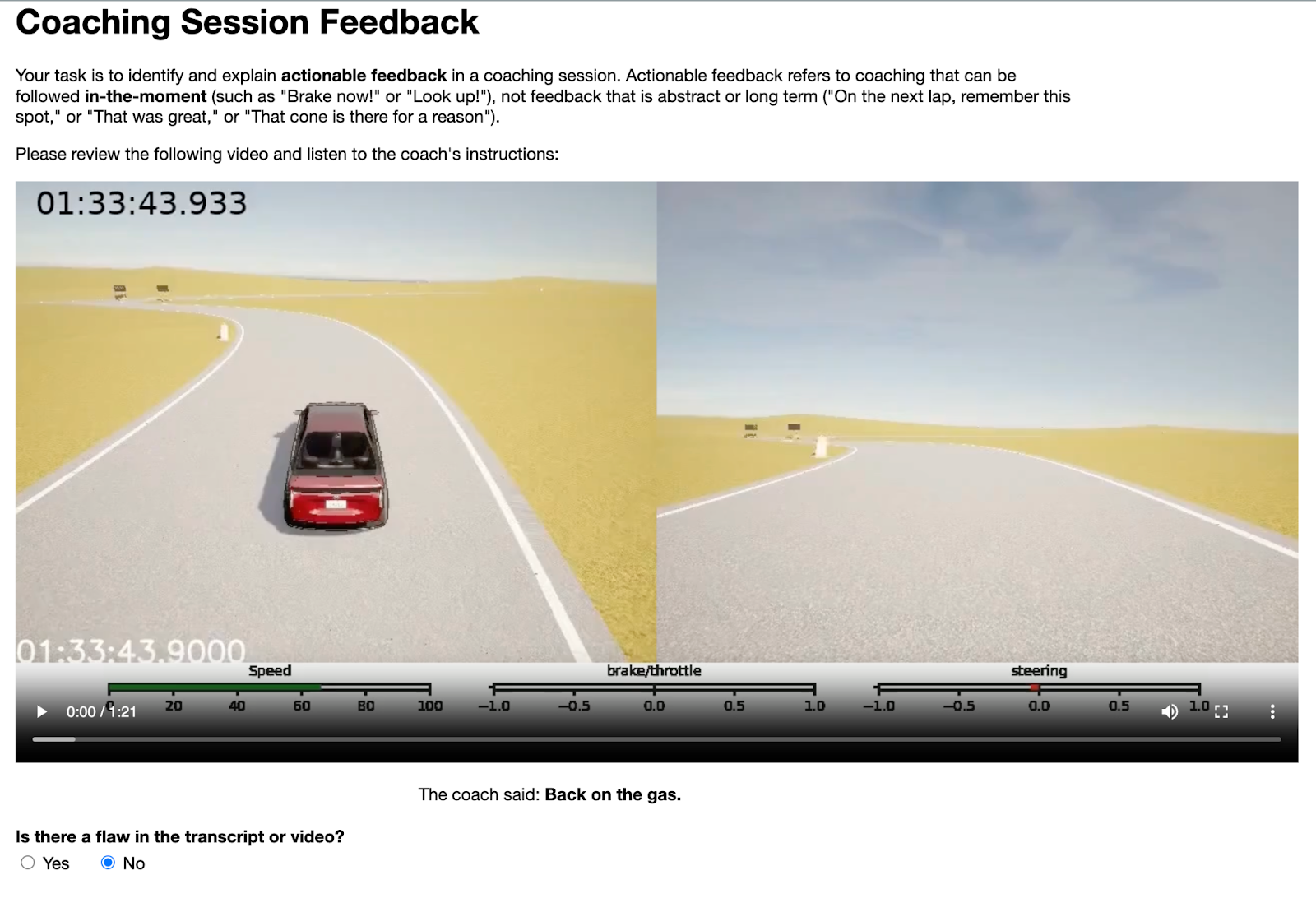}
    \end{minipage}
    \begin{minipage}{0.4\textwidth}
        \centering
        \includegraphics[width=1.0\linewidth]{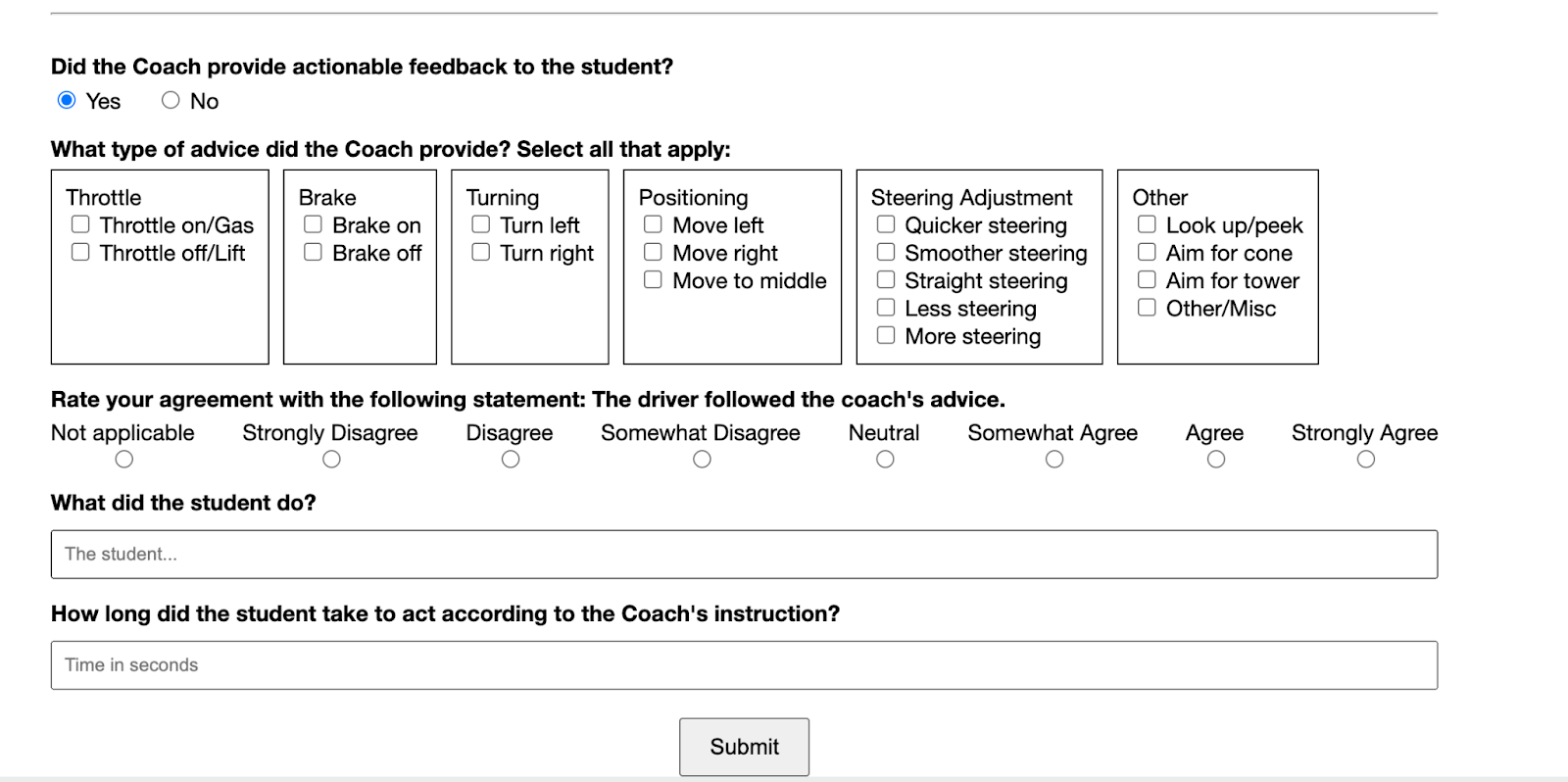}
    \end{minipage}
    \begin{minipage}{0.4\textwidth}
        \centering
        \includegraphics[width=1.0\linewidth]{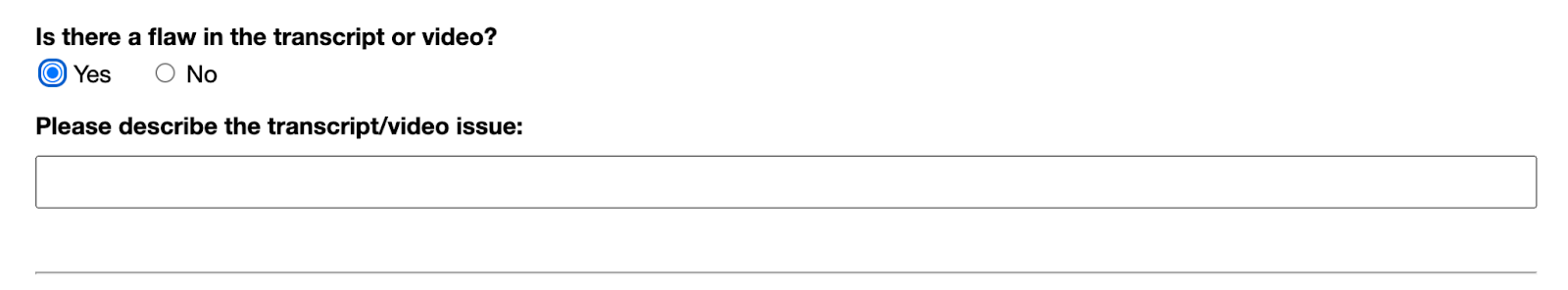}
    \end{minipage}
    
    \caption{The annotation interface used for compliance annotations.}
\end{figure}

Annotators were shown a video of how to complete the task. 

Here, we will go over each question:
\begin{enumerate}
    \item \textbf{Did the coach provide actionable feedback to the student?} 
    The coach often provides feedback that the student cannot act on, saying things like:
        \begin{itemize}
            \item Good job!
        \end{itemize}
        \begin{itemize}
            \item You were late there.
        \end{itemize}
        \begin{itemize}
            \item Remember this turn on your next lap
        \end{itemize}
        \begin{itemize}
            \item How was that?
        \end{itemize}

    If the coach’s advice is \textbf{not actionable}, please respond \textbf{``No''} to this question, and skip the remaining questions.

    \item \textbf{Is there a flaw in the transcript or video?}
    Our transcripts might not always be right, or the timing might be incorrect. While it should be rare, please respond \textbf{Yes} if there are any noticeable flaws in the transcript (for example, incorrect timing or content). If you respond with \textbf{Yes}, a small text-box will pop-up asking you to describe the error.
     Please write a short description of what you think is wrong (for example, “The transcript is wrong, the coach didn’t say that.” or “The text in the transcript does not appear in the video” or “There is no transcript” or “The video looks corrupted”)

    \item \textbf{What type of advice did the Coach provide? Select all that apply.}
    For this question, we have provided several ``themes'' of advice, such as ``Throttle on'', ``Throttle off'', ``Move left”, “Smoother steering”, etc. Please choose the category (or categories) that best aligns with the coach’s advice. Some things to consider:
        \begin{itemize}
            \item You will hear ``stay right'' or ``move to the left'', note that these are \textbf{different} from ``turn right'' or ``turn left''. We have different categories for ``move'' and ``turn''.
        \end{itemize}
        \begin{itemize}
            \item Annotators were all provided with several examples
        \end{itemize}

    \item \textbf{How well did the student follow the Coach's advice?}
    For this question, please use your best judgment based on the video (what the car did and how the driver controls changed) to rate the student’s compliance with the coach. There is a ``Not Applicable'' field for advice that you cannot verify, such as ``Look up'', ``Look for the cone'', ``Look right'', etc.

    \item \textbf{What did the student do?}
    Please write out a complete sentence describing the driver’s behavior. For example:
        \begin{itemize}
            \item The driver accelerated while turning left.
        \end{itemize}
        \begin{itemize}
            \item The student moved to the right side of the track.
        \end{itemize}
        \begin{itemize}
            \item The student spun out on the grass.
        \end{itemize}
        \begin{itemize}
            \item The driver lifted off of the gas to slow down.
        \end{itemize}

    \item \textbf{How long did the student take to act according to the Coach’s instruction?} 
    For this question, please use the video slider to find the amount of time between when the coach finished speaking and the student started to comply with the advice. We call this the student’s reaction time. Please enter a number in this box, reflecting the student’s reaction time (in seconds). If \textbf{reaction time} is not applicable (such as a “look” command, or if the driver never listens to the coach), simply respond with -1.        
\end{enumerate}

    \subsection{FAQs}
    \begin{itemize}
        \item \textbf{For what duration of time should we monitor the student to ensure they are following the coach's guidance after the coach gave the instruction? One example is: Coach says ``Turn to left” for the future event (after 4 / 5 seconds) then asks the driver to turn right and then turn left. So, in this case, the response time would be around 8 seconds.}
        For exceptionally ambiguous examples, please flag such videos with the ``Is there a flaw in the transcript/video” radio button, and we will review them. We have added a text box so that it is possible to explain why something is wrong, such as if a transcript is incorrect or unclear.

        As a general rule, we only want you to consider the first utterance in the video (also represented as text beneath the video). Any other instructions that are given during the video should be ignored. In the above case, it sounds like the driver did not comply with the first instruction.

        \item \textbf{Could you clarify if the instruction given by the coach "Find the cone/tower" is equivalent to ``Aim for the cone/tower"? As of now, we have assumed it to be equivalent and marked it as actionable.}
        Find the cone/tower should be labeled as ``Look” advice. So switch to N/A for compliance and -1 for reaction time.

        \item \textbf{If there is a slight turn on the track and the coach gives instructions as just ``left” or ``right”, should it be categorized as a ``Turn" or as ``Positioning"? We are categorizing it in positioning as of now.}
        Positioning is correct.

        \item \textbf{If the coach gives the same instruction more than once then which instruction should we consider for calculation of reaction time? (as of now we are considering the first one) For Example : The coach says ``Turn, Turn, Turn”.}

        \item \textbf{If the student is already following the instruction such as staying right on the track, what should be the reaction time? We have marked a few of these examples with 0.1 seconds reaction time.}
        Please mark any data that is already happening as a 0.0 reaction time.

        \item \textbf{Should ``Stay on the gas" be categorized as ``Throttle on" , if so how should the rating be given for the same as for most of the cases the student keeps the same gas (neither increases or decreases)? Should it be ``Strongly agree"?}
        Yes, consider this is ``throttle on”, and set as ``strongly agree” if the student seems to be following the advice very well. The coach may be saying ``stay in the gas" to tell student to stop slowing down, so it might mean ``apply more throttle". The advice is contextual, so do your best to interpret what is being conveyed in each utterance.

        \item \textbf{How should we evaluate a student's performance when instructions are ``Squeeze", ``Back on the gas" or ``Throttle On"? Should the rating be based on the amount of throttle applied (e.g., less than half throttle = ``Somewhat agree," more than half = ``Agree," and full throttle = ``Strongly agree") or based on student's reaction (e.g., if student applied half throttle then it should be ``Strongly agree")?}
        That is a reasonable stance to take. I would say that some of the time, it is contextual (the coach might want half throttle, not full throttle), but in general using the amount of throttle is a good proxy for amount of satisfaction/compliance.

    \end{itemize}


\section{Annotation guidelines for sentence categorization annotation task}\label{appendix:cf-annotation-guidelines}

This guide provides guidelines for completing the sentence categorization annotation task successfully.

The guide contains:

\begin{enumerate}
    \item Specification of the annotation task.
    \item Guidelines for training the annotators and general things to keep in mind
    \item Specific failure cases that have happened during the initial test phases and ways in which to rectify the mistakes.
\end{enumerate}

\subsection{Annotation task specification}
The following prompt was shown to the annotator. 
\begin{quote}
    The sentence shown on the screen was said by a car racing coach during a sim-racing coaching session. These sentences are short phrases. Your task is to annotate the sentence with a tuple of comma-separated values (Type, Category, Subcategory). The possible tuples can be seen in the drop-down menu. Additionally, for each phrase there will be a (type, category, subcategory) already shown next to it. You need to correct the tuple if it is wrong. 
\end{quote}

\subsection{General guidelines for annotator training}
As part of the annotator training procedure the annotators were required to study expert annotations of utterances from eight different trials. These annotations were created by the researchers and the experimenters in consultation with the coach and served as a template for the annotators for annotating the full set of utterances. The following guidelines were given to the annotators as they studied the expert examples. 
\begin{quote}
Before you start your annotation job, please get very familiar with the examples provided at ground truth examples. These examples cover a wide variety of sentences and their corresponding correct ground truth tuples. Learn how (and why) different sentences are mapped to the corresponding tuples. Feel free to refer back to these examples while you are doing the annotation (if anything confuses you). Strive your best to make an informed decision. Please be aware of mouse slips when choosing the options from the drop down menu. If you are completely unsure about which one to choose, please select the option `Unsure', and leave a reason why you are unsure in the text box.
\end{quote}

\subsection{Specific failure cases and error mitigation strategies}

Based on some of the initial tests that the annotators took, we observed there are few types of commonly recurring errors. We provided additional guidelines to minimize those errors. Here are some examples from the guidelines. 

\begin{enumerate}
    \item “You got it”, “Then right”
    \begin{enumerate}
        \item Ground truth: "You got it" (feedback, positive, none), "Then Right" (instruction, right, stay) respectively
        \item Annotation Error: (Instruction Throttle On), (feedback, positive, none) respectively
        \item Possible reason for errors: This is likely because the annotator may have gotten distracted and hurried through the task. Or it could have been a mouse slip. In either case, \textbf{please make sure you take enough time to think it through before selecting the correct option}
    \end{enumerate}
    \item “Stay in it”
    \begin{enumerate}
        \item Ground truth: (Instruction, throttle, \textbf{stay})
        \item Annotation Errors: (Instruction Throttle On), (Commentary Driving Recommendation)
        \item Mitigation: Pay attention to the verb in the phrase. If it is “stay” then the subcategory will be stay, if it is “move” then the subcategory will be move. Since the phrase is fairly short, this is of type “instruction” and not “commentary”. In general, think of whether this phrase is uttered by the coach so that they want the student driver to act according to the phrase immediately or later. If immediately, then the phrase is to be annotated as an instruction.
    \end{enumerate}
    \item ``Aim for the tower"
    \begin{enumerate}
        \item Ground truth: (instruction, steering, landmark)
        \item Annotation Errors: (instruction, looking ahead, landmark.)
        \item Mitigation: Since the ‘verb’ is ‘aim for”, we will have ‘steering’ as the category. If the verb was “over to”, the category would be ‘steering’. If the verb was ‘look for’, or “find” then the category would be ‘looking ahead’. 
    \end{enumerate}
    \item I need you to turn a little earlier, no brakes on this right-hand turn, okay?, you can still be a little quicker
    \begin{enumerate}
        \item Ground truth: (commentary, driving, recommendation) respectively
        \item Annotation Errors: (instruction, turn, none), (instruction, brake, off), (instruction, steering, quickness) respectively
        \item Mitigation: Key is to look at the length of the sentence and make a judgment regarding whether the coach is expecting the student to act upon the phrase immediately or not. Similar to the mitigation strategy in 2c.
    \end{enumerate}
\end{enumerate}

\section{Annotation guidelines terminal feedback sentence categorization}\label{appendix:tf-annotation-guide}
One annotator went through the terminal feedback and annotated instruction categories using MAXQDA \cite{maxqda_2021-tm}. Figure~\ref{fig:maxqda} shows the interface the annotator used. They followed these instructions:

\begin{figure}[H]
    \centering
    \includegraphics[width=0.8\linewidth]{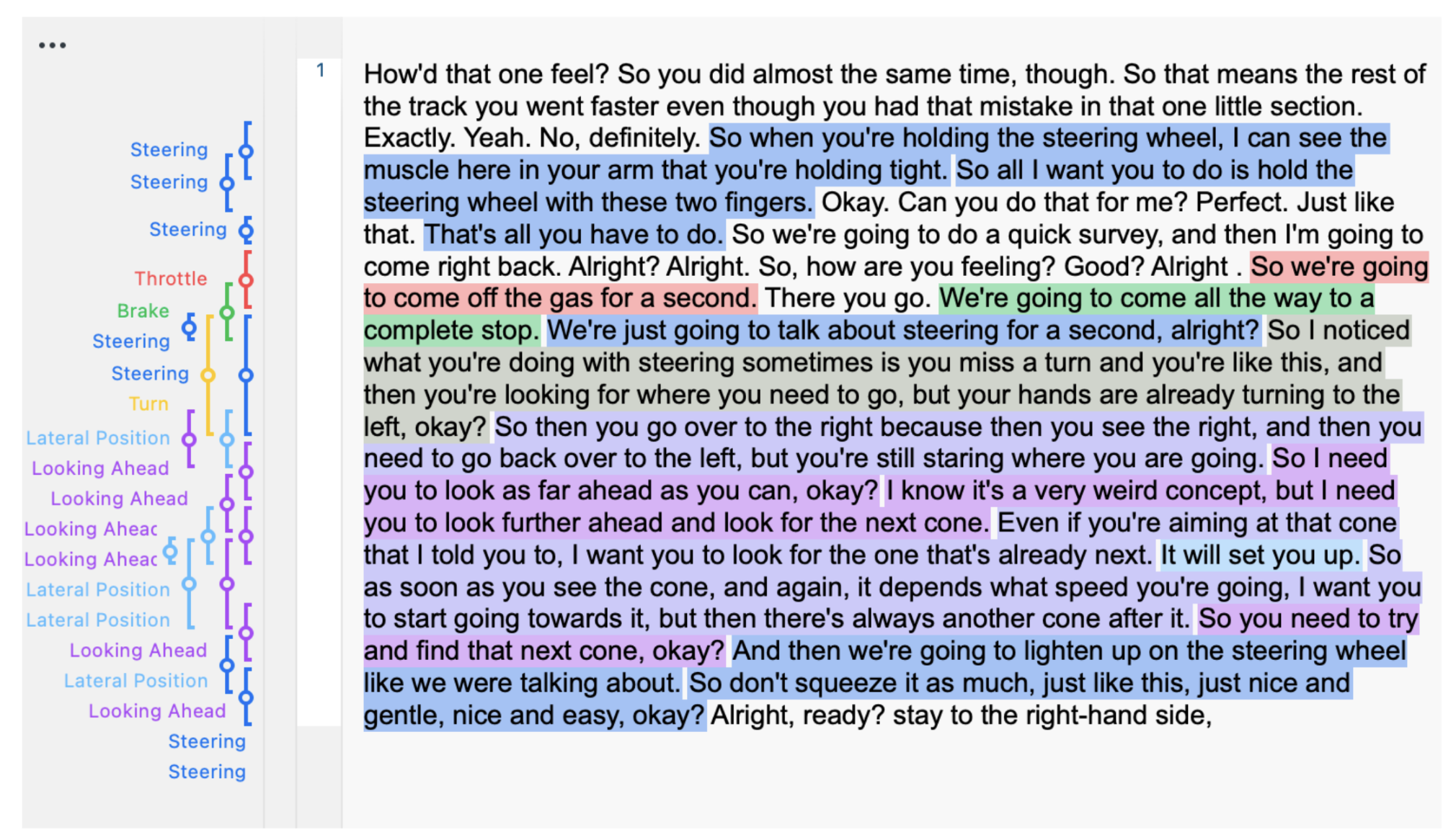}
    \caption{MAXQDA -- the coding interface the annotator used for terminal feedback annotation.}
    \label{fig:maxqda}
\end{figure}

\begin{enumerate}
    \item Upload one participant at a time's worth of data to MAXQDA  (note you have to use the xls format for uploading data into MAXQDA)
    \item Go through each terminal feedback cell, Select a category for each sentence where one of the main topics is present. Both implicit AND explicit mentions of each category. If there is no meaningful instruction-related terminal feedback in the whole dialogue, code with ``No meaningful terminal feedback''
    \item Add a column to each of the terminal feedback transcripts for the main categories – add counts (brake, looking ahead, lateral position, throttle, steering, turn, other (some conversation related to instruction but not in one of the above categories), no terminal feedback on instruction (only use if the whole dialogue has no instruction-related terminal feedback)
    \item Export codes per participant

\end{enumerate}

Throttle is instructions related to how much gas is being applied to the throttle (e.g., `more gas'). Lateral position has to do with instructions related to where the car is positioned on the track (e.g., `move to the right'). Looking ahead instruction has to do with where the student is looking (e.g., `look at the cone'). Braking is the instructions related to how much pressure to apply to the brakes and when (e.g., `brake now'). Turning is about explaining when and where to turn  (e.g., `now turn in'). Steering is related to moving the car towards visual aids (e.g., `move closer to that cone').

\section{Time syncing and cleaning the coach and participants' audio transcripts}
This is a guide on how to properly verify and edit the audio transcripts. This guide is specifically for Subtitle Composer. Please note that this system only works for Linux/Ubuntu machines (apt-get should work for package subtitle composer also for Ubuntu 18.04). 

\subsection{Getting Started}
\begin{enumerate}
    \item Download the desired srt and mp3 files
    \item Go to files and open your driver/student.srt and coach.srt using Subtitle Composer. 
        \begin{itemize}
            \item This should open 2 different windows.
        \end{itemize}
    \item Split your screen in half with the driver.srt on one side and coach.srt on the other. 
        \begin{itemize}
            \item For the picture below, the driver is on the right and the coach is on the left.
            \begin{figure}[H]
                \centering
                \includegraphics[width=0.7\linewidth]{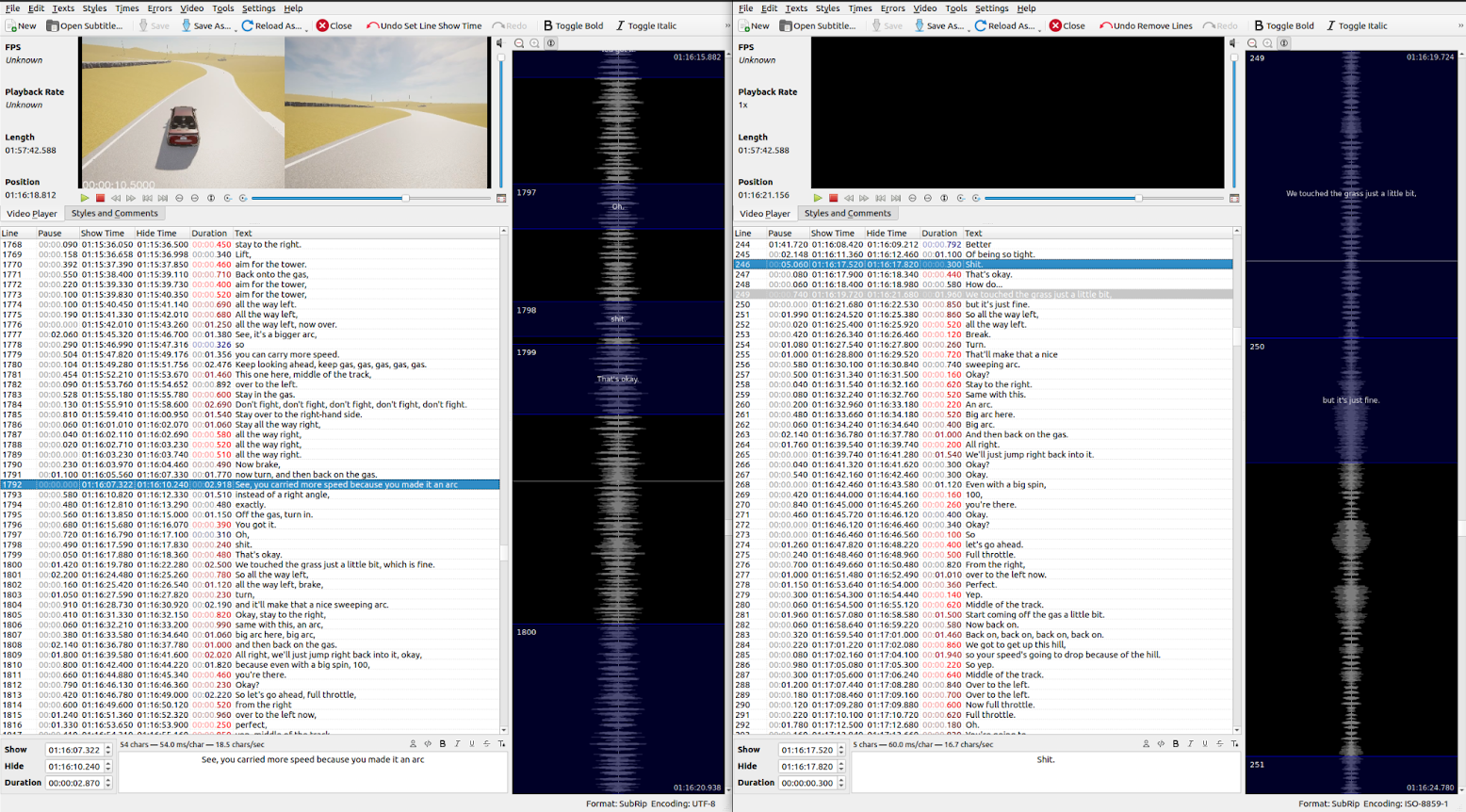}
                \caption{Two Subtitle Composer windows.}
                \label{fig:enter-label}
            \end{figure}  
    \end{itemize}
\end{enumerate}

\subsection{Editing the Transcripts}
There are \textbf{5} major edits that need to be made for these transcripts.

\begin{enumerate}
    \item \textbf{Timing}:
    Often, there can be times when the subtitle is in the wrong place or does not encapsulate the audio it represents. Click on the subtitle in the audio track and move the subtitle to the desired place. You can also make the subtitle longer or shorter by dragging the top and bottom to the appropriate place. And then save your edits.

    Another way to fix timing is if there are sentences broken up in weird places, or fast repeating phrases, you can select those sentences and combine them. To do this, select the subtitles you want to combine, right-click and press ``Join Lines''. In the text box, edit the newly combined subtitles. Then save your edits.

    See the dictionary in Table~\ref{tab:subtitle-dictionary} below for more information on combining and separating sentences. 
        \begin{figure}[H]
            \centering
            
            \begin{minipage}{0.45\linewidth}
                            \centering
                \includegraphics[width=0.5\linewidth]{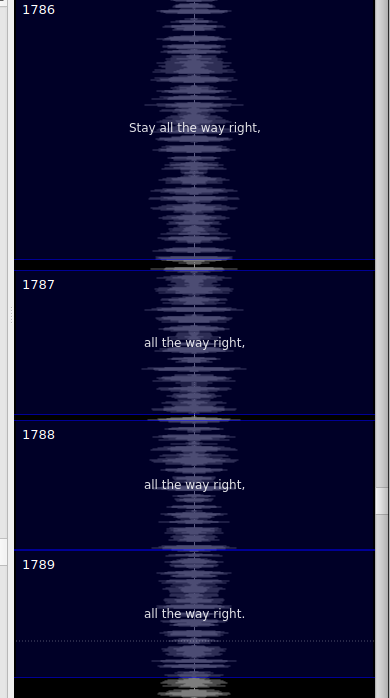}

                (a)
            \end{minipage}
            \begin{minipage}{0.45\linewidth}
                \centering
\vspace{1.25cm}
                
                \includegraphics[width=\linewidth]{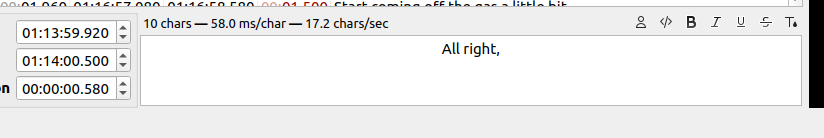}

\vspace{1.35cm}
                (b)
            \end{minipage}
            \caption{(a) Subtitle Composer Audio Window (b) Text editing box. }
            \label{fig:enter-label}
        \end{figure}
    \item \textbf{Spelling/grammar errors}
    While listening to the audio, correct any spelling mistakes you see in the text box and then save your edits.
    The most common errors that you will see are ``all right'' and ``break''. 
    There is a dictionary below of more common mistakes. Please add any repeated mistakes you have seen below.
    \item \textbf{Missing words/sentences}
    Please ensure that you are listening to the audio files in full, as the transcription can miss what the speaker is saying. If this happens, right-click in the audio track and insert a new line. While on the new line, type in the missed information that the transcription missed. After fixing that error, save your edits. 
    \item \textbf{Cross talk}
    The .srt files are divided between the driver and coach. If you are listening to the coach audio file and editing that .srt, you want to delete any driver dialogue that the transcript picked up. This will be the same for the driver as well. After deleting, save your edits.	
    \item \textbf{Proctor talk}
    For high-quality transcripts, you want to make sure you are only tracking the voice of the driver or the coach. Therefore, any directions or conversations from/with the study proctor must be removed. Select the subtitles that have those conversations and delete them from the file. Then, save your work. 
\end{enumerate}

%



\begin{table}[h]
\caption{Common mistakes annotators were told to look out for.}
\label{tab:subtitle-dictionary}
\centering
\small
\begin{tabularx}{\textwidth}{|p{3.3cm}|X|X|}
\hline

\textbf{Mistake Type} & \textbf{Mistake} & \textbf{Correction} \\
\hline
Spelling/Grammar & ``All right'' & ``Alright'' \\
\hline
Spelling/Grammar & ``Break'' (slowing down) & ``Brake'' \\
\hline
Spelling/Grammar & ``Stay on the gas'' & ``Stay in the gas'' — more likely what the coach is saying. Listen carefully. \\
\hline
Spelling/Grammar & ``Soaring at the steering wheel'' & ``Sawing at the steering wheel'' \\
\hline
Separating long sentences & ``stay to the left hand side aim for the tower now over to the right hand side'' & Separate into: ``stay to the left hand side'', ``aim for the tower'', ``now over to the right hand side''. Split according to distinct instructions for accurate timing. \\
\hline
Separating repeated sentences & ``Stay to the right, stay to the right, stay to the right'' & If full phrases are repeated, break them into individual subtitles. \\
\hline
Combining sentences & ``Stay'' \newline ``To the left'' & Combine into one command — they go together. \\
\hline
\end{tabularx}
\end{table}

\section{Questionnaires}\label{appendix:questionnaires}
\subsection{Standard surveys}
\subsubsection{NASA TLX}
Participants answered each question on a scale of 1 -21:
        \begin{itemize}
            \item How mentally demanding was the task?
            \item How physically demanding was the task?
            \item How hurried or hushed was the pace of the task?
            \item How successful were you in accomplishing what you were asked to do?
            \item How insecure, discouraged, irritated, stressed, and annoyed were you?
        \end{itemize}

\subsubsection{PANAS short form}
\begin{itemize}
        \item \textbf{PANAS Positive:} Please rate the extent to which you feel the following at the present moment on the scale (Very Slightly or not at all, A little, Moderately, Quite a bit, Very much)
        \begin{itemize}
            \item Determined
            \item Attentive
            \item Alert
            \item Inspired
            \item Active
        \end{itemize}
        \item \textbf{PANAS Negative:} Please rate the extent to which you feel the following at the present moment on the scale (Very Slightly or not at all, A little, Moderately, Quite a bit, Very much)
        \begin{itemize}
            \item Afraid
            \item Nervous
            \item Upset
            \item Ashamed
            \item Hostile
        \end{itemize}        
\end{itemize}

\subsubsection{Intrinsic motivation inventory}
For each of the following statements, please indicate how true it is for you on a 7 point Likert scale.
    \begin{itemize}
        \item While I was working on the task I was thinking about how much I enjoyed it.
        \item I did not feel at all nervous about doing the task.
        \item I felt that it was my choice to do the task.
        \item I think I am pretty good at this task.
        \item I found the task very interesting.
        \item I felt tense while doing the task.
        \item I think I did pretty well at this activity, compared to other students.
        \item Doing the task was fun.
        \item I felt relaxed while doing the task.
        \item I enjoyed doing the task very much.
        \item I didn't really have a choice about doing the task.
        \item I am satisfied with my performance at this task.
        \item I was anxious while doing the task.
        \item I thought the task was very boring.
        \item I felt like I was doing what I wanted to do while I was working on the task.
        \item I felt pretty skilled at this task.
        \item I thought the task was very interesting.
        \item I felt pressured while doing the task.
        \item I felt like I had to do the task.
        \item I would describe the task as very enjoyable.
        \item I did the task because I had no choice.
        \item After working at this task for awhile, I felt pretty competent.
    \end{itemize}

\subsection{All participants}
\subsubsection{Student pre-drive survey}
\textbf{Driving Type:} The scale for questions 1-7 is the following: Everyday, A few times a week, A few times a month, A few times a year, About Once a year, Never.
\begin{enumerate}
    \item How often do you use your own vehicle to do the following activities within the span of a year? Select one per row.
    \begin{itemize}
        \item Commuting
        \item Driving friends/family
        \item Running errands
        \item Longer distance trips 
    \end{itemize}
    \item How often do you use your own vehicle to do the following activities within the span of a year? Select one per row. 
        \begin{itemize}
            \item Towing/hauling
            \item Recreation/motor sport
            \item Ride sharing (driving for Lyft or Uber)
            \item Off-roading
        \end{itemize}
    \item How frequently do you use your own vehicle for trips of the following distances?
        \begin{itemize}
            \item Trips of less than 100 miles
            \item Trips between 100 and 199 miles
            \item Trips between 200 and 299 miles
            \item Trips of 300 or more miles
        \end{itemize}
    \item On average, how many miles a week do you drive? (please enter a number - if you do not drive enter 0)
    \item How would you describe your experience with video games?
        \begin{itemize}
            \item I do not play video games
            \item I rarely play video games (less than once a month)
            \item Casual Gamer (I play video games occasionally)
            \item Regular Gamer (I play video games regularly)
            \item Hardcore Gamer (I dedicate significant time to gaming)
        \end{itemize}
    \item How would you describe your experience with IRacing?
        \begin{itemize}
            \item I do not use iRacing
            \item I used iRacing once
            \item I used iRacing A few times
            \item I use IRacing monthly
            \item I use iRacing weekly
        \end{itemize}
    \item How would you describe your experience with Gran Turismo?
          \begin{itemize}
            \item I do not use Gran Turismo
            \item I used Gran Turismo
            \item I used Gran Turismo A few times
            \item I use Gran Turismo monthly
            \item I use Gran Turismo weekly
        \end{itemize}  
    \item How would you rate your level of experience with karting?
          \begin{itemize}
            \item Novice (I have minimal or no experience with karting)
            \item Intermediate (I have some experience with recreational karting or have attended a karting school)
            \item Advanced (I regularly participate in karting events, such as local leagues or club races)
            \item Expert (I have competed in regional, national, or international karting competitions)
        \end{itemize} 
    \item How would you rate your level of experience with performance driving?
          \begin{itemize}
            \item Novice (I have minimal or no experience with performance driving)
            \item Intermediate (I have some experience with performance driving, including a few track days or performance driving courses)
            \item Advanced (I regularly participate in performance driving events, such as track days, racing schools, or competitive events)
            \item Expert (I have extensive experience in performance driving and have participated in numerous competitive events or professional racing)
        \end{itemize} 
    \item I-PANAS-SF
\end{enumerate}

\subsubsection{Student mid-drive survey}
\begin{enumerate}
    \item Do you feel like you are progressing?
        \begin{itemize}
            \item Yes
            \item No
        \end{itemize}
    \item On a scale from 1 - 10, how much fun are you having? (1 - no fun at all. 10 - the most fun I've ever had)
    \item NASA-TLX
    \item I-PANAS-SF
\end{enumerate}

\subsubsection{Student post-drive survey}
\begin{enumerate}
    \item How much potential do you think you have for sim racing?
        \begin{itemize}
            \item No potential
            \item A little potential
            \item A moderate amount of potential
            \item A lot of potential
            \item A great deal of potential
        \end{itemize}
    \item On a scale from 1 - 10, how excited would you be to do this again? (1 - not excited at all. 10 - very excited)
    \item Intrinsic Motivation Inventory    
\end{enumerate}

\subsection{Coached participants only}
\subsubsection{Familiarization survey}
\begin{enumerate}
    \item On a scale from 1 - 10, how much fun are you having? (1 - no fun at all. 10 - the most fun I've ever had)
    \item On a scale from 1 - 10, how comfortable are you operating the driving simulator? (1 - not comfortable at all, 10 - extremely comfortable)
    \item I-PANAS-SF
    \item NASA-TLX (on a scale of 1 -21)
\end{enumerate}

\subsubsection{Student mid-drive survey}
For the coaching group, the following question was added to this survey.
\begin{enumerate}
    \item On a scale from 1 - 10, how helpful are the instructions you are receiving from your coach? (1 - not helpful at all. 10 - very helpful)
\end{enumerate}

\subsubsection{Student post-drive survey}
For the coaching group, the following question was added to this survey.
\begin{enumerate}
    \item On a scale from 1 - 10, how helpful are the instructions you are receiving from your coach? (1 - not helpful at all. 10 - very helpful)   
\end{enumerate}

\subsection{Coach surveys} 
\subsubsection{Coach familiarization survey}
\begin{enumerate}
    \item What is your initial assessment of the student's driving skill? (1 being novice, 10 being highly skilled)
    \item What aspect do you think the student needs to work on the most?
        \begin{itemize}
            \item Steering
            \item Throttle
            \item Braking
            \item Looking Ahead
        \end{itemize}
    \item On a scale of 1-to-10, 1 being novice and 10 being an expert, how would you rate this student's...
        \begin{itemize}
            \item Steering
            \item Throttle
            \item Braking
            \item Looking Ahead
        \end{itemize}
    \item On a scale of 1-to-10, 1 being very uncomfortable and 10 being very comfortable, how would you rate the student's comfort driving the simulator?
\end{enumerate}

\subsubsection{Coach mid-drive survey}
\begin{enumerate}
    \item Is the student progressing?
        \begin{itemize}
            \item Yes
            \item No
        \end{itemize}
    \item What aspects did you work on during your last session? (Multiple answers allowed)
        \begin{itemize}
            \item Steering
            \item Throttle
            \item Braking
            \item Looking Ahead
        \end{itemize}
    \item What aspect do you think the student needs to work on the most
         \begin{itemize}
            \item Steering
            \item Throttle
            \item Braking
            \item Looking Ahead
        \end{itemize}   
    \item On a scale of 1-to-10, 1 being very uncomfortable and 10 being very comfortable, how would you rate the student's comfort driving the simulator?
    \item How would you rate the student's overall skill at the moment (1 being novice, 10 being expert)
\end{enumerate}

\subsubsection{Coach post-drive survey}
\begin{figure}[H]
    \centering
    \includegraphics[width=.95\linewidth]{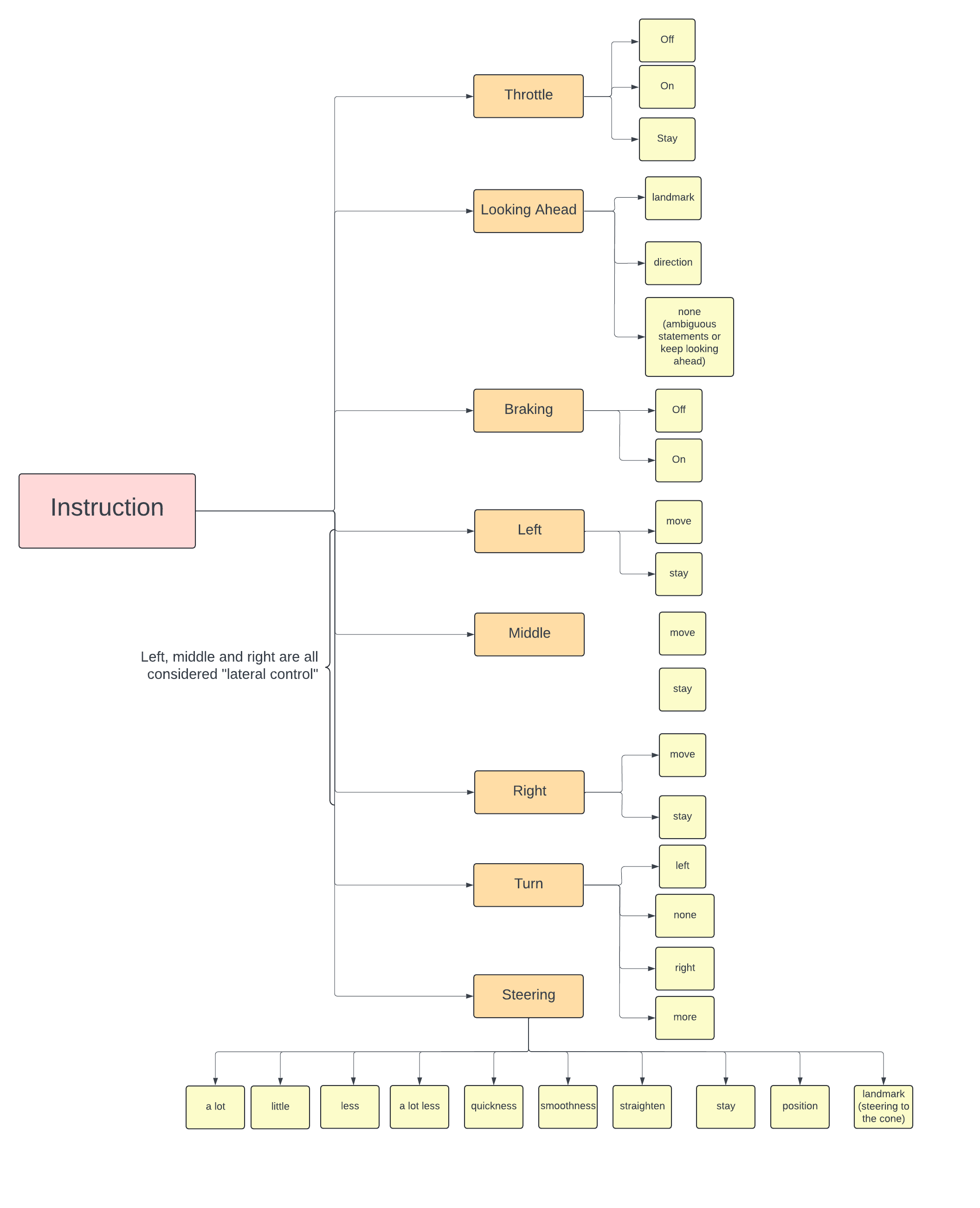}
    \caption{Overview of the instructions taxonomy, as a two-level hierarchy.}
    \label{fig:taxonomy-instruction}
\end{figure}

\begin{figure}[H]
    \centering
    \includegraphics[width=0.96\linewidth]{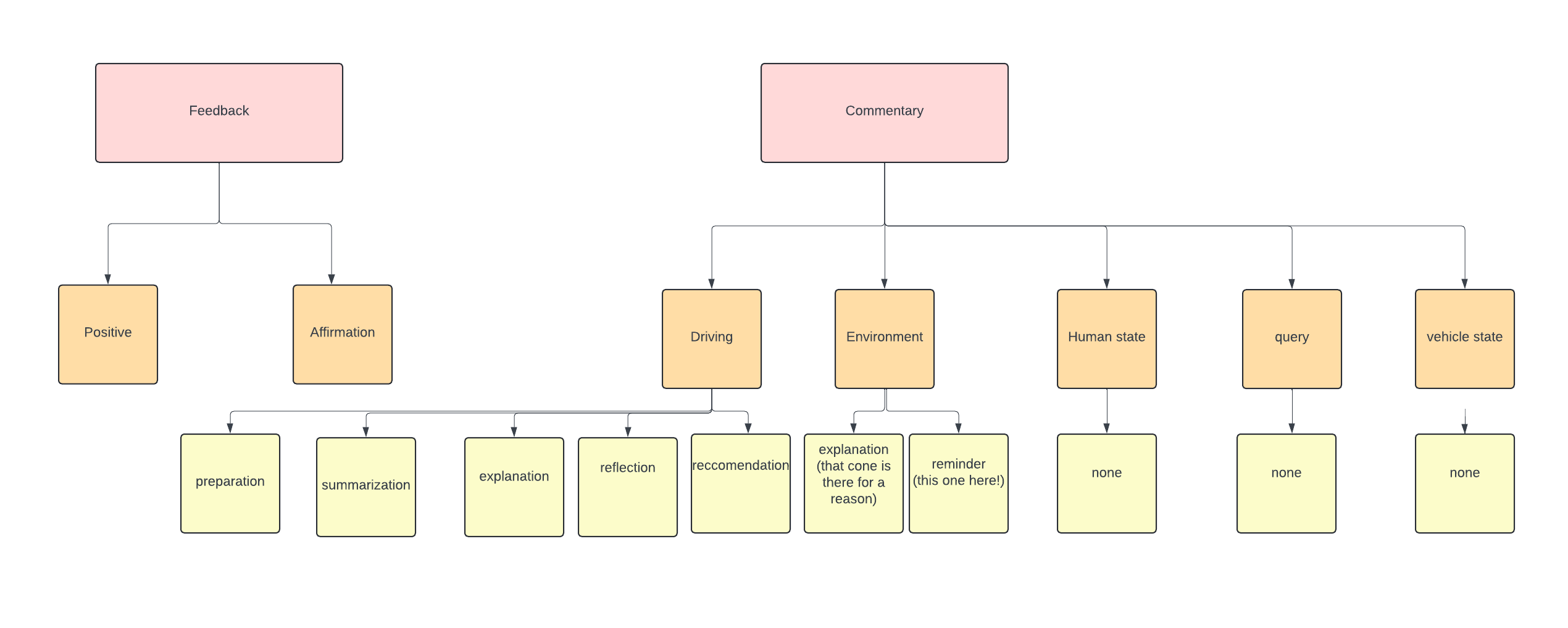}
    \caption{Overview of the feedback and commentary taxonomy, as a two-level hierarchy }
    \label{fig:taxonomy-label}
\end{figure}
\begin{enumerate}
    \item What is your final assessment of the student's driving skill? (1 being novice, 10 being expert)
    \item What aspect do you think the student needs to work on the most
         \begin{itemize}
            \item Steering
            \item Throttle
            \item Braking
            \item Looking Ahead
        \end{itemize} 
    \item On a scale of 1-to-10, 1 being novice and 10 being an expert, how would you rate this student's...
         \begin{itemize}
            \item Steering
            \item Throttle
            \item Braking
            \item Looking Ahead
        \end{itemize} 
    \item On a scale of 1-to-10, 1 being very uncomfortable and 10 being very comfortable, how would you rate the student's comfort driving the simulator?
    \item How receptive was the student to coaching, 1 being not receptive and 10 being extremely receptive
    \item How motivated to become a better driver was the student (1 being not motivated at all, 10 being extremely motivated)
    
\end{enumerate}

\section{Taxonomy labels} \label{appendix:tax-details}
Figures~\ref{fig:taxonomy-instruction} and \ref{fig:taxonomy-label} describe the taxonomy used for annotations. The three high-level categories are instruction, commentary, and feedback. This taxonomy was developed through qualitative interviews with the coach, as well as through reviewing the transcripts. 

\section{Track map}
\begin{figure}[H]
    \centering
    \includegraphics[width=0.9\linewidth]{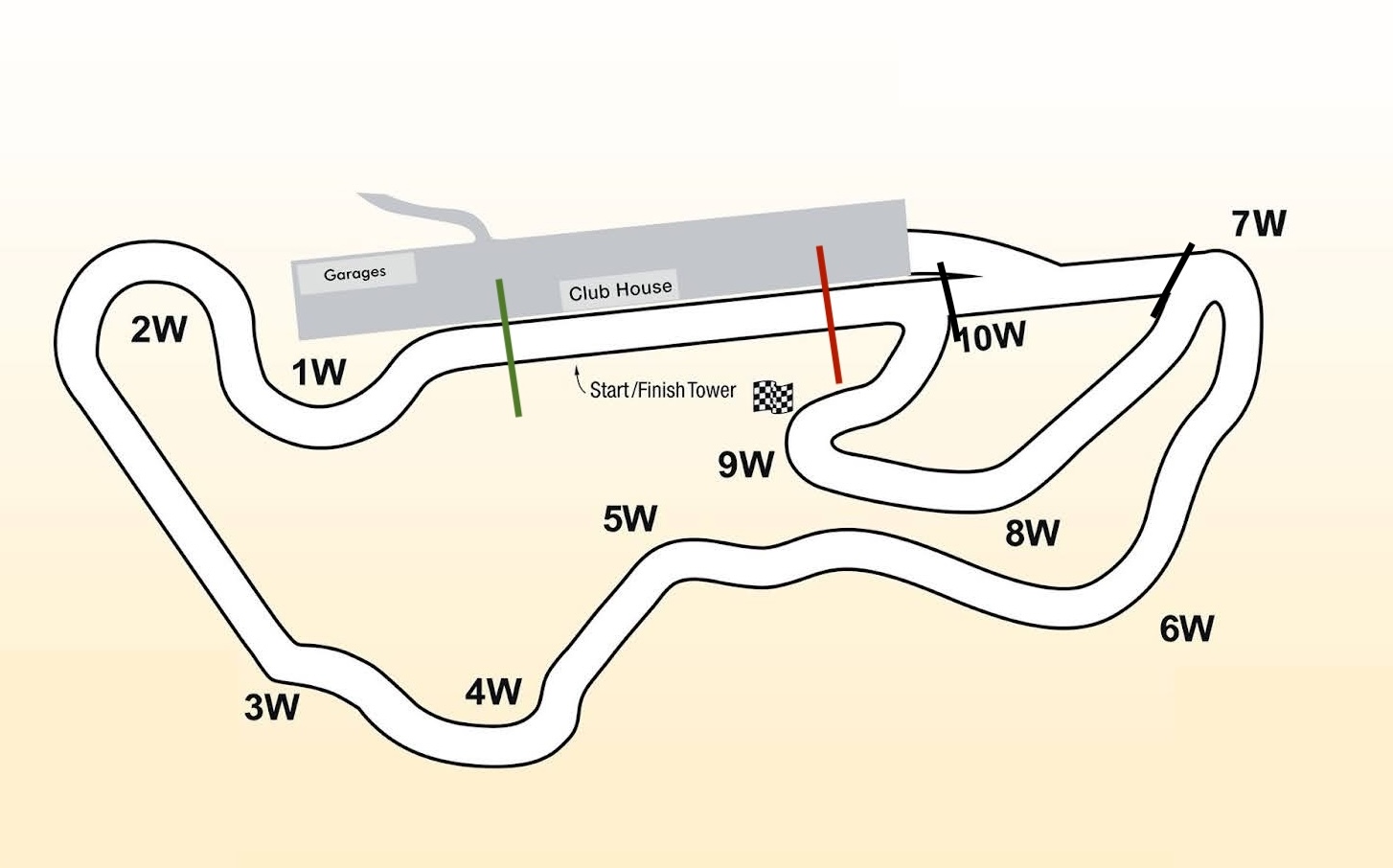}
    \caption{Thunderhill Raceway Track with start (green) and end (red) gates marked.}
    \label{fig:trackmap}
\end{figure}

Figure~\ref{fig:trackmap} contains the track map for the Thunderhill Raceway Track, with marked start and end gates.

\end{document}